\documentclass{article}
\PassOptionsToPackage{numbers, compress}{natbib}

\usepackage[preprint]{neurips_2026}

\usepackage[utf8]{inputenc} 
\usepackage[T1]{fontenc}    
\usepackage{hyperref}       
\usepackage{url}            
\usepackage{booktabs}       
\usepackage{amsfonts}       
\usepackage{nicefrac}       
\usepackage{microtype}      
\usepackage{xcolor}         

\usepackage{amsmath}
\usepackage{graphicx}
\usepackage{caption}   
\usepackage{subcaption} 
\usepackage{wrapfig}
\usepackage{algorithm}
\usepackage{listings}
\usepackage{multirow}
\usepackage{tabularx}

\usepackage[table]{xcolor}

\newif\ifdraft
\drafttrue

\ifdraft

\newcommand{\hadar}[1]{{\color{magenta}[\textbf{Hadar:} #1]}}

\newcommand{\noam}[1]{{\color{cyan}[\textbf{Noam:} #1]}}

\newcommand{\sagie}[1]{{\color{green}[\textbf{Sagie:} #1]}}

\else
\newcommand{\hadar}[1]{}
\newcommand{\noam}[1]{}
\newcommand{\sagie}[1]{}

\fi

\title{Colored Noise Diffusion Sampling}

\author{%
  Hadar Davidson \quad
  Noam Issachar \quad
  Sagie Benaim \\
  \\
  The Hebrew University of Jerusalem
}

\begin{document}

\maketitle

\newcommand{\fix}{\marginpar{FIX}}
\newcommand{\new}{\marginpar{NEW}}

\begin{abstract}
Diffusion models achieve state-of-the-art image synthesis, with their generative trajectories fundamentally exhibiting a spectral bias, resolving low-frequency global structures early and high-frequency fine details later. Conventional stochastic differential equation (SDE) solvers fail to account for this dynamic, naively injecting uniform white noise throughout the entire process and misusing the finite energy budget. In this work, we establish a mathematical framework that reconsiders SDE inference as a targeted, frequency-decoupled energy transfer. Leveraging this framework, we introduce Colored Noise Sampling (CNS), a novel, training-free stochastic solver. Rather than injecting uniform white noise, CNS utilizes a dynamic, timestep- and frequency-dependent schedule that more efficiently allocates injected energy toward structurally unresolved frequency bands. By actively exploiting the model's inherent spectral bias, CNS systematically steers the generated distribution toward the true data manifold. 
Extensive experiments demonstrate that CNS significantly outperforms standard ODE and SDE baselines as a strictly plug-and-play, inference-time sampler substitution across diverse architectures (SiT, JiT, FLUX). Compared to standard sampling on ImageNet-256, CNS achieves substantial unguided FID reductions, improving from 8.26 to 6.27 on SiT-XL/2, 32.39 to 26.69 on JiT-B/16, and 11.88 to 8.31 on JiT-H/16, while yielding consistent relative FID improvements with Classifier-Free Guidance. Project page is available at \url{https://hadardavidson.github.io/CNS/}.
\end{abstract}
\section{Introduction}    
\label{sec:intro}

Diffusion models have established a new standard in photorealistic image synthesis, defining the state-of-the-art in high-fidelity generation \citep{ho2020denoising, song2020score, flux-2-2025}. 
Crucially, the sampling trajectory of these models exhibits a \textit{spectral bias} \citep{wang2023diffusion, issachar2025dype}. This inductive property dictates that diffusion models inherently resolve low-frequency global structures during early sampling steps, while filling in high-frequency fine details in later steps. 

Current sampling algorithms \citep{song2019generative, song2020score, song2020denoising} fail to account for this phenomenon. Standard stochastic methods based on Stochastic Differential Equations (SDEs) naively inject uniform white noise, completely disregarding how the frequency spectrum of the generated image dynamically evolves over time. To address this inefficiency at its root, we introduce a new class of stochastic solvers that actively leverages this spectral bias. By tailoring the injected noise to the specific denoising timestep, we improve generation fidelity without requiring any additional training.

Recognizing the importance of spectral bias, several recent works have attempted to exploit it. One line of research alters the training framework, interpolating from spectrally non-uniform or temporally evolving noise distributions \citep{falck2025fourier, scimeca2025learning, huang2024blue}. Other approaches operate at inference time, introducing ad-hoc modifications such as frequency-decoupled operations \citep{qian2024boosting, yu2026elucidating}, internal activation reweighting \citep{si2024freeu}, or step-size schedule adjustments \citep{lee2025beta}. While these methods yield measurable improvements, they remain fundamentally constrained by their underlying use of spectrally uniform solvers. This naturally leads to our guiding question: \textit{How can we actively exploit the spectral bias of diffusion models to design a fundamentally new, general-purpose sampler that improves generation fidelity?}

\begin{figure}[t]
  \centering
  \setlength{\tabcolsep}{0pt} 
  \renewcommand{\arraystretch}{0} 
  
  \makebox[\textwidth][c]{
    \begin{tabular}{@{}c @{\hspace{5pt}} cccccc c@{}} 
      
      \raisebox{0.08\textwidth}{\rotatebox[origin=c]{90}{\footnotesize \textbf{ODE}}} &
      \includegraphics[width=0.16\textwidth]{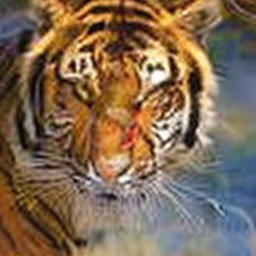} &
      \includegraphics[width=0.16\textwidth]{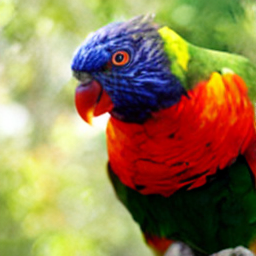} &
      \includegraphics[width=0.16\textwidth]{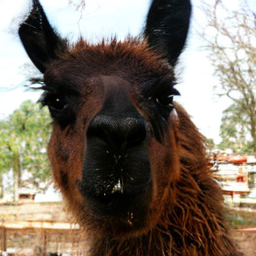} &
      \includegraphics[width=0.16\textwidth]{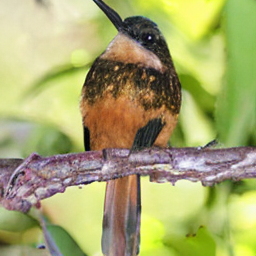} &
      \includegraphics[width=0.16\textwidth]{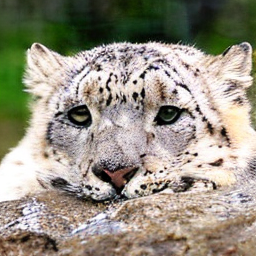} &
      \includegraphics[width=0.16\textwidth]{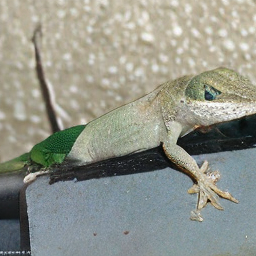} &
      \phantom{\rotatebox[origin=c]{90}{\scriptsize \textbf{CNS (Ours)}}} \\[-12pt] 
      
      \raisebox{0.08\textwidth}{\rotatebox[origin=c]{90}{\footnotesize \textbf{SDE}}} &
      \includegraphics[width=0.16\textwidth]{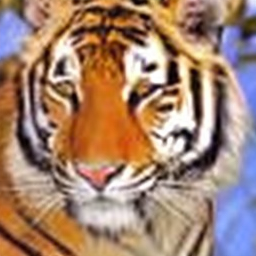} &
      \includegraphics[width=0.16\textwidth]{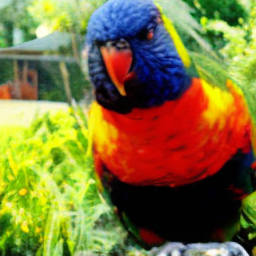} &
      \includegraphics[width=0.16\textwidth]{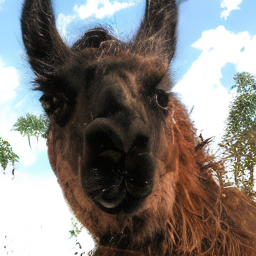} &
      \includegraphics[width=0.16\textwidth]{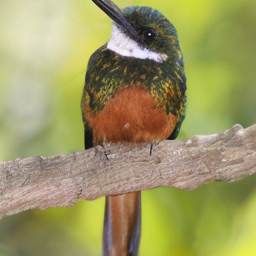} &
      \includegraphics[width=0.16\textwidth]{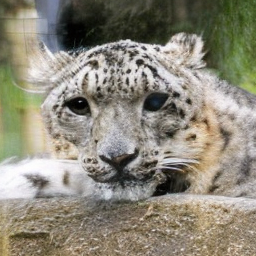} &
      \includegraphics[width=0.16\textwidth]{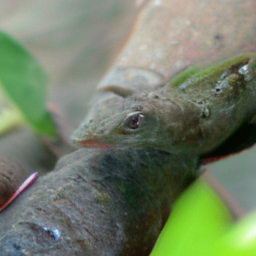} &
      \phantom{\rotatebox[origin=c]{90}{\scriptsize \textbf{CNS (Ours)}}} \\[-12pt]
      
      \raisebox{0.08\textwidth}{\rotatebox[origin=c]{90}{\footnotesize \textbf{CNS (Ours)}}} &
      \includegraphics[width=0.16\textwidth]{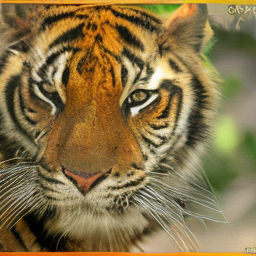} &
      \includegraphics[width=0.16\textwidth]{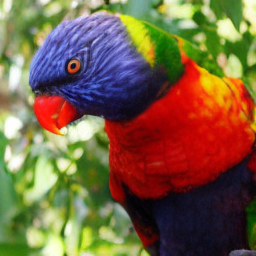} &
      \includegraphics[width=0.16\textwidth]{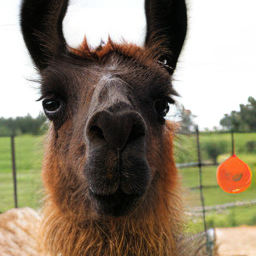} &
      \includegraphics[width=0.16\textwidth]{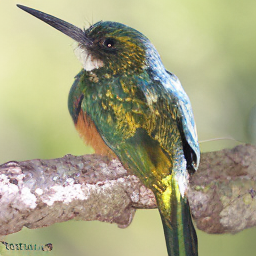} &
      \includegraphics[width=0.16\textwidth]{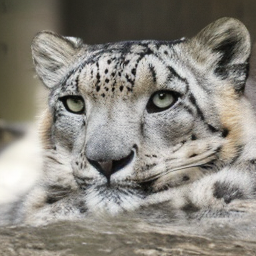} &
      \includegraphics[width=0.16\textwidth]{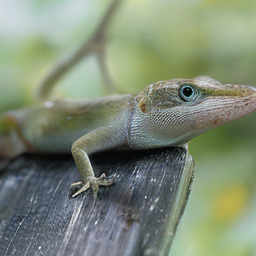} &
      \phantom{\rotatebox[origin=c]{90}{\scriptsize \textbf{CNS (Ours)}}} \\
    \end{tabular}
  } 
  \vspace{-0.2cm}
  
  \caption{\textbf{Colored Noise Sampling (CNS)}. Samples from SiT-XL/2 on ImageNet-256 (with CFG) for different sampling strategies. While standard SDEs inject uniform white noise, our Colored Noise Sampling (CNS) dynamically reallocates injected stochastic energy to unresolved frequency bands. This actively leverages the network's spectral bias to systematically steer the output toward the true data manifold, outperforming standard ODE and SDE solvers.}
  \vspace{-0.5cm}
  \label{fig:teaser}
\end{figure}

To answer this, we first establish a mathematical framework to control the generated distribution via frequency-aware noise injection. Geometrically, sampling trajectories resemble non-orthogonal rotations toward the data manifold \citep{wang2023diffusion}. This implies that diffusion models do not arbitrarily discard initial noise; rather, a significant structural component of this signal is preserved and mapped into final image features \citep{xu2025good, mao2023guided}.

A key observation of our work is that this signal-preserving transfer also applies to the continuous noise injected by SDE solvers throughout the trajectory. Furthermore, this process is frequency-decoupled: injected noise in a specific frequency band maps directly to spatial features in that same band. By ensuring our frequency-aware adjustments remain strictly variance-preserving, requiring only that the total injected energy per step remains normalized, we demonstrate that the classic Langevin requirement for uniform white noise \citep{oksendal2003stochastic} can be safely relaxed without pushing intermediate states out of distribution.

Building on this framework, we construct a timestep- and frequency-dependent noise schedule. By analyzing the progression rates of different frequency bands during generation, we propose that the network's ability to convert injected noise into coherent image features significantly depends on how structurally ``resolved'' that specific band is at a given timestep. This insight allows us to reconsider SDE sampling as a targeted \textit{energy injection} process. Rather than uniformly distributing the finite injected noise budget, our approach utilizes a dynamic schedule based on the expected evolution of the trajectory, allocating energy to the frequency bands where it is most needed. This principled allocation steers the output toward the true data manifold, yielding strictly higher-fidelity generation.

To validate our approach, we conduct extensive experiments across diverse architectures and modalities, including latent-space generation (SiT \citep{ma2024sit}), pixel-space generation (JiT \citep{li2025back}), and state-of-the-art text-to-image synthesis (FLUX \citep{flux2024, flux-2-2025}). Evaluated primarily via the Fréchet Inception Distance (FID) \citep{heusel2017gans}, empirical results demonstrate that our method significantly outperforms standard ODE and SDE baselines. On ImageNet-256 \citep{russakovsky2015imagenet}, we achieve substantial FID reductions under both unguided and Classifier-Free Guidance (CFG) \citep{ho2022classifier} settings, while maintaining robust stability across varying discretization steps. We visually highlight the superiority of our approach over standard baselines in Fig.~\ref{fig:teaser}. Furthermore, our sampler proves effective when integrated into text-to-image pipelines like FLUX, improving automatic human-preference scores.

To summarize, our main contributions are: 

\begin{itemize}
    \item We establish a mathematical framework that reframes SDE noise injection as a targeted energy transfer, and demonstrate that the standard Langevin requirement for spectrally uniform white noise can be safely relaxed to resolve spectral gaps.
    
    \item We introduce Colored Noise Sampling (CNS), a novel, training-free stochastic solver that actively leverages spectral bias by dynamically allocating injected noise energy toward structurally unresolved frequency bands.

    \item We validate CNS as a robust, general-purpose sampler across diverse architectures (SiT, JiT, FLUX). On ImageNet-256, CNS achieves substantial unguided FID reductions (e.g., 8.26 to 6.27 for SiT-XL/2, 32.39 to 26.69 for JiT-B/16) and relative CFG improvements ranging from $\sim$8\% to $\sim$50\% over standard ODE and SDE baselines.
\end{itemize}

\section{Related Work}
\label{sec:related_work}

\noindent \textbf{Samplers for Diffusion Models.} \quad 
\label{subsec:related_sampling}
Sampling in diffusion models is a highly researched domain primarily focused on numerically mitigating discretization errors. Prominent advancements include higher-order solvers \citep{wu2024stochastic, xue2023sa, lu2022dpm} that maintain fidelity at low step counts, dynamic solver alternation \citep{liu2024unified, zou2025usf++}, and state reparameterizations that smooth integration pathways \citep{zheng2023dpm, zhang2022fast, chen2025tada}. While these methods successfully reduce truncation errors and accelerate generation, they remain agnostic to the evolving spatial structure of the state. Our approach is fundamentally orthogonal: rather than strictly optimizing numerical precision, we optimize the allocation of stochastic energy by explicitly exploiting the model's spectral bias. 

\noindent \textbf{Leveraging Spectral Bias in Diffusion Models.} \quad
A prominent line of research exploits the spectral bias of diffusion models during training by altering noise distributions. These methods rely on empirical heuristics to modify initial \cite{scimeca2025learning} or temporally-evolving noise distributions \cite{huang2024blue}, or introduce formally grounded frequency-dependent processes like EqualSNR \cite{falck2025fourier}. However, fundamentally altering the learning objective demands costly model retraining. In sharp contrast, our approach overcomes this barrier by introducing a purely plug-and-play sampler that harnesses spectral bias exclusively at inference time.
To circumvent retraining costs, a separate line of work leverages spectral bias via inference-only modifications. These methods introduce ad-hoc adjustments to the generation pipeline, such as applying frequency-decoupled operations to the predicted state \citep{qian2024boosting, yu2026elucidating}, dynamically reweighting internal network activations \citep{si2024freeu}, adjusting step-size schedules \citep{lee2025beta}, or coupling spectral bias with positional encodings \cite{issachar2025dype}. While effective, these techniques treat the underlying stochastic solver as a static black box. Our work targets this unexplored component: rather than modifying the network or its outputs post-hoc, we directly embed the spectral bias into the core sampling mechanism itself.

\section{Method}
\label{sec:method}
We now introduce our approach. Sec.~\ref{sec:preliminaries} outlines standard diffusion background. Sec.~\ref{sec:spectral_bias} and \ref{sec:energy_transfer} formalize the specific generative phenomena, inference-time spectral bias and noise energy preservation, that we leverage to build our framework. Building on these principles, Sec.~\ref{sec:method_generated_distribution} analyzes the spectral gap induced by standard SDEs. Finally, Sec.~\ref{sec:method_CNS} details how CNS dynamically colors injected noise to actively steer the generated spectrum toward the true data manifold.

\subsection{Background: Diffusion Models and Sampling Dynamics}
\label{sec:preliminaries}
Diffusion Models \citep{ho2020denoising, song2020score} and Flow Matching \citep{lipman2022flow, liu2022flow} can be unified under the continuous-time framework of Stochastic Interpolants \citep{albergo2022building}.
Given a target data distribution $x_0 \sim p_{\text{data}}$ and a tractable noise prior $\epsilon \sim \mathcal{N}(0, I)$, these models construct a probability path via the time-dependent state:
\begin{equation}
    x_t = \alpha_t x_0 + \sigma_t \epsilon ,\quad t \in [0, 1]
\end{equation}
Boundary conditions are established such that $x_0$ strictly represents the clean data ($\alpha_0=1, \sigma_0=0$) and $x_1$ approximates the pure noise prior ($\alpha_1 \approx 0, \sigma_1 \approx 1$). Whether utilizing trigonometric schedules like Variance-Preserving (VP) diffusion or linear paths like Flow Matching ($\alpha_t = 1-t, \sigma_t = t$), the objective remains learning to reverse this continuous-time probability flow.

To learn this reverse flow, these models approximate the conditional interpolant velocity:
\begin{equation}
    v_t = \dot{\alpha}_t x_0 + \dot{\sigma}_t \epsilon
\end{equation}
Because the intermediate state $x_t$ is a simple affine combination of data and noise, predicting the velocity $v_\theta$ is algebraically equivalent to predicting the clean data $x_\theta \approx x_0$, the noise $\epsilon_\theta \approx \epsilon$, or the marginal score $s_\theta \approx \nabla_{x_t} \log p_t(x_t)$. Omitting explicit $(x_t, t)$ dependencies for brevity, these parameterizations are deterministically linked by the following relations:
\begin{equation}
    v_\theta = \dot{\alpha}_t x_\theta + \dot{\sigma}_t \epsilon_\theta, \quad x_t = \alpha_t x_\theta + \sigma_t \epsilon_\theta, \quad s_\theta = -\frac{\epsilon_\theta}{\sigma_t}
\end{equation}
During inference, novel samples are generated from the prior by substituting these learned predictions into reverse-time differential equations, which are then integrated using either deterministic or stochastic solvers.

\noindent \textbf{Sampling Dynamics.} \quad Deterministic sampling formulates the trajectory as a Probability Flow ODE (PF-ODE), directly integrating the predicted velocity:
\begin{equation}
    dx_t = v_\theta(x_t, t) dt, \quad x_0 = x_1 + \int_{1}^{0} dx_t \approx x_1 + \sum_{i} v_\theta(x_{t_i}, t_i) \Delta t_i
\end{equation}
While computationally efficient and approximately invertible, this strict determinism lacks an inherent corrective mechanism. Consequently, discrete numerical approximations and network errors inevitably accumulate, causing intermediate states to gradually drift off the true data manifold and degrade final image fidelity \citep{song2020denoising}.

Stochastic solvers address this drift by simulating the generative process as a reverse-time SDE. Introducing a time-dependent diffusion coefficient $g(t) > 0$ and a reverse-time Wiener process $\bar{\mathrm{w}}$, the dynamics expand to:
\begin{equation}
    dx_t = \left( v_\theta(x_t, t) - \frac{1}{2}g(t)^2 s_\theta(x_t, t) \right) dt + g(t) d\bar{\mathrm{w}}
\end{equation}
\begin{wrapfigure}{r}{0.45\textwidth}
  \vspace{-1em}
  \centering
  \includegraphics[width=0.95\linewidth]{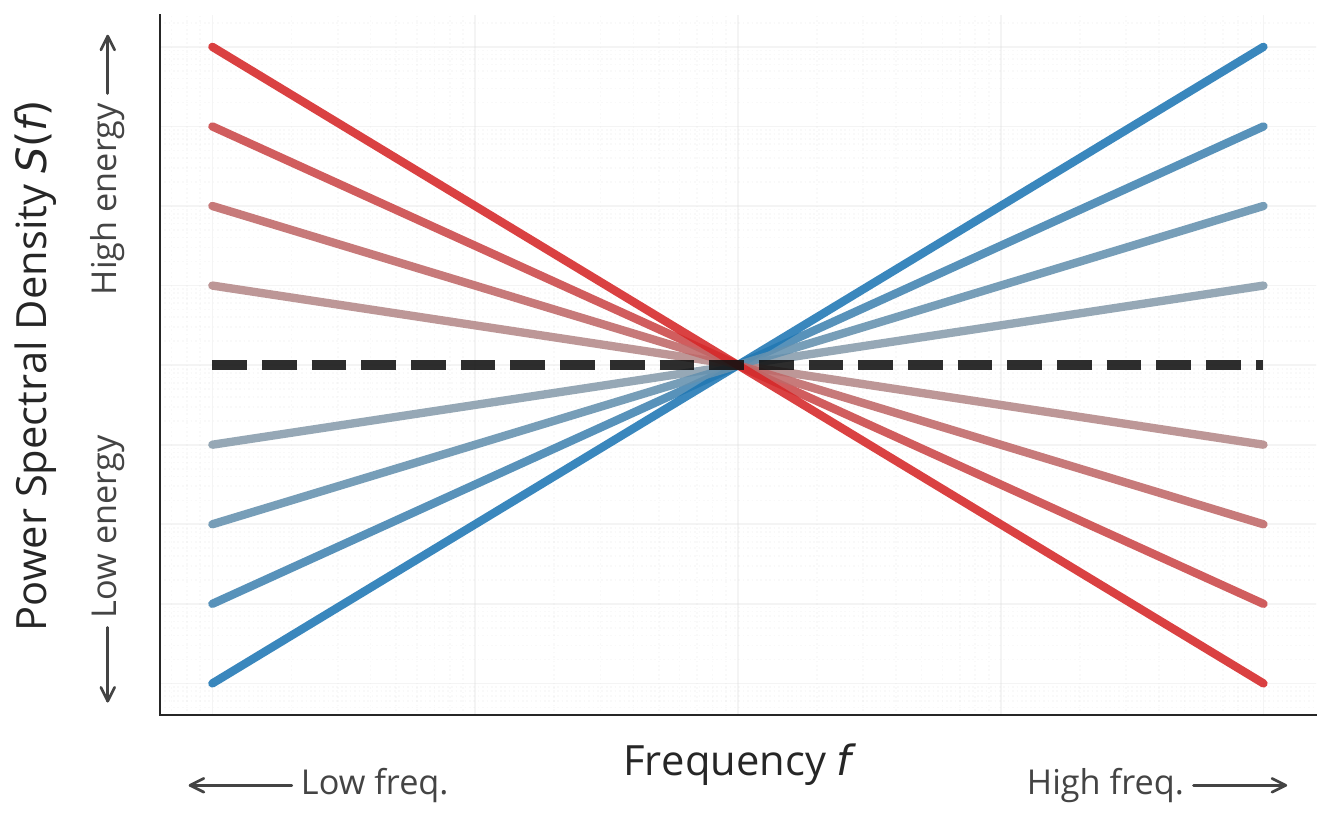}
  \caption{PSD of different colored noises. The spectra transition smoothly from high-frequency dominant \textcolor{blue}{blue noise}, through uniform white noise (center black line), to low-frequency dominant \textcolor{red}{red noise}.}
  \label{fig:psd_noise_colors}
  \vspace{1em}
\end{wrapfigure}
This process fundamentally alters the trajectory by continuously counterbalancing white Gaussian noise injection with a restorative gradient step along the predicted score. The injected noise explores the local latent neighborhood, while the score-based denoising actively pulls the state back toward high-density regions. By natively correcting accumulated discretization errors at every step, SDE solvers keep the trajectory firmly anchored to the true data distribution, yielding superior visual quality \citep{song2019generative, song2020score}.

\noindent \textbf{Power Spectral Density and Noise Colors.} \quad The frequency composition of the injected noise $\epsilon$ is characterized by its Power Spectral Density (PSD). 
Letting $\hat{\epsilon}(f) = \mathcal{F}(\epsilon)$ denote its Fourier transform, the PSD $S(f)$ evaluates the expected energy at frequency $f$: 
\begin{equation}
S(f) = \mathbb{E}[|\hat{\epsilon}(f)|^2]
\end{equation}

The shape of $S(f)$ defines the noise ``color'' \citep{oppenheim1999discrete} as illustrated in Fig.~\ref{fig:psd_noise_colors}. Standard Gaussian noise $\epsilon \sim \mathcal{N}(0, I)$ possesses a constant $S(f)$, injecting equal energy across all frequencies (\textit{white noise}). Conversely, non-uniform spectra produce colored noise, such as high-frequency dominant \textit{blue noise}. Due to Fourier's orthogonality, Parseval's theorem ensures that integrating the PSD yields the total spatial energy: $\mathbb{E}[\|\epsilon\|^2] = \int S(f) df$. Consequently, standard SDEs fundamentally operate by blindly injecting a fixed, frequency-agnostic white noise energy budget at every generative step.

\subsection{Spectral Bias of Diffusion Models}
\label{sec:spectral_bias}
Spectral bias is a well-documented inductive property that extends beyond training optimization to fundamentally govern the inference dynamics of diffusion models \citep{rahaman2019spectral, ronen2019convergence, wang2023diffusion}. Rather than resolving the image uniformly, generation follows a staggered frequency evolution.

To formalize this band-wise progression, we evaluate the model's clean data prediction at each intermediate timestep $t$. Under a linear schedule, this prediction is given by:
\begin{equation}
    x_{t}^{pred} = x_\theta(x_t,t)= x_t -t\,v_\theta(x_t, t)
\end{equation}
Let $X_0(f) = \mathcal{F}(x_0)(f)$ and $X^{pred}(f,t) =\mathcal{F}(x_{t}^{pred})(f)$ denote the spectral components at frequency band $f$ for the final generated latent and the intermediate prediction, respectively. Following \citep{issachar2025dype}, we measure the resolved energy of this intermediate prediction relative to the final outcome to define the bounded progress index $\gamma(f, t) \in [0, 1]$ for every frequency band $f$:
\begin{wrapfigure}{r}{0.35\textwidth}
  \vspace{-0.6em}
  \centering
  \includegraphics[width=0.99\linewidth]{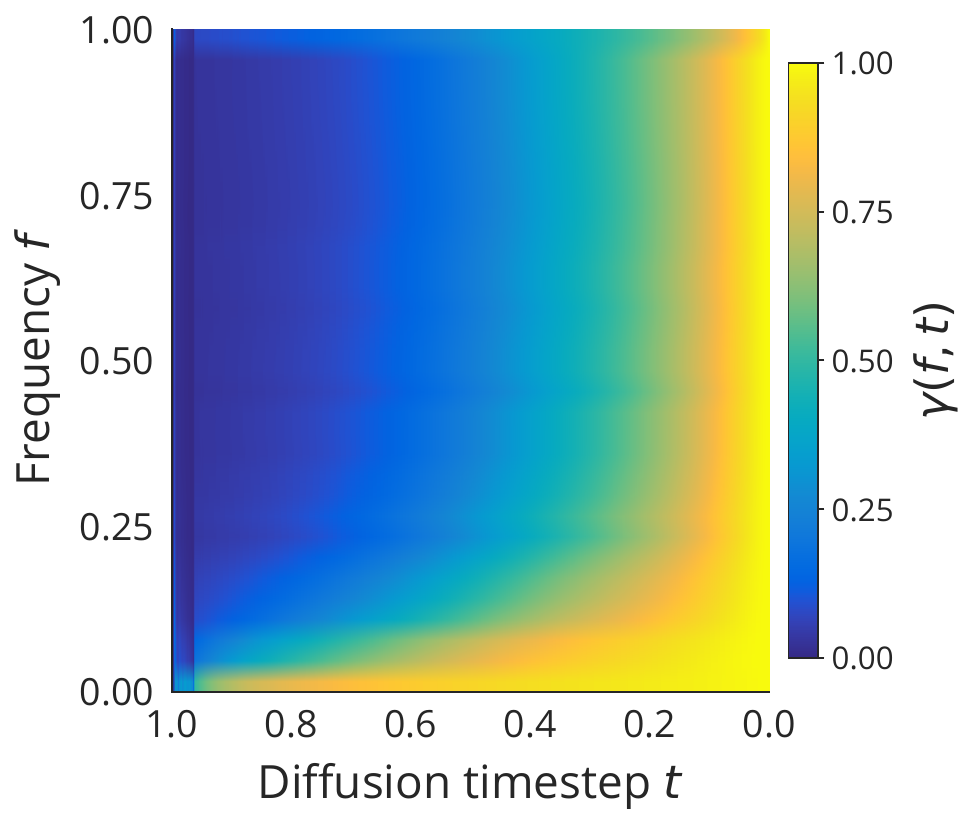}
  \caption{Temporal progression of frequency bands during sampling.}
  \label{fig:gamma_matrix}
  \vspace{-2.5em}
\end{wrapfigure}
\begin{equation}
    \gamma(f, t) = 1 - \frac{|X_0(f) - X^{pred}(f,t)|^2}{|X_0(f)|^2}
\end{equation}
This index isolates exactly how much of a specific frequency band's final structure has been resolved by the network at any given timestep $t$ (see Alg.~\ref{alg:gamma_mat_calc} and App.~\ref{app:subsec:radial_frequency_usage} for further details). Visualizing this $\gamma$-matrix (Fig.~\ref{fig:gamma_matrix}) directly exposes these generation dynamics: low-frequency structures resolve early in the generation process. In contrast, high-frequency details evolve at a gradual rate, only fully materializing at the very end of the sampling trajectory. Ultimately, this provides a precise temporal map dictating exactly when specific frequency bands are actively being ``built'' by the network.

\subsection{Structural Preservation and Energy Transfer in Diffusion Models}
\label{sec:energy_transfer}

The mapping from the prior $\mathcal{N}(0, I)$ to the data distribution $p_{\text{data}}$ is not an arbitrary coupling between the two spaces. Empirical evidence demonstrates that the inference process preserves significant information from the initial noise realization \citep{yan2025beyond, xu2025good, staniszewski2024there}, naturally following minimal-distance trajectories \citep{issachar2025designing}.

To explain this geometrically, \citet{wang2023diffusion} demonstrate that sampling trajectories are surprisingly low-dimensional. Rather than taking an unconstrained walk across the latent space, these trajectories effectively resemble 2D rotations of $\theta \approx 1$ radian from the initial noise state toward the target data manifold. In high-dimensional spaces, where independent random vectors are nearly orthogonal, this rotational angle yields a remarkably high expected cosine similarity ($\cos(\theta) \gg 0$). This mathematically confirms that the diffusion process does not generate novel structures from scratch, but rather preserves a substantial portion of the initial structural signal---a phenomenon we empirically visualize across spatial frequencies in Fig.~\ref{fig:cos_sim_ODE_SDE_freq_Dependent}.

This rotational perspective has profound implications for the sampling dynamics. 
Since rotations preserve the $L_2$ norm (i.e., the vector's energy), inference acts as a signal transfer mechanism that retains a significant portion of the initial noise's energy.
The model deterministically maps this preserved noise onto the structured spatial features of the final image, a property implicitly leveraged by recent works optimizing initial noise selection \citep{zhou2025golden, ahn2024noise, mao2023guided}. This non-destructive property forms the foundational premise of our approach: it implies that by strategically controlling the energy injected into the process via noise, we directly control the structural features of the final image.

\subsection{The Generated Image Distribution Spectrum}
\label{sec:method_generated_distribution}

\begin{figure}[t]
  \centering  \includegraphics[width=0.93\linewidth]{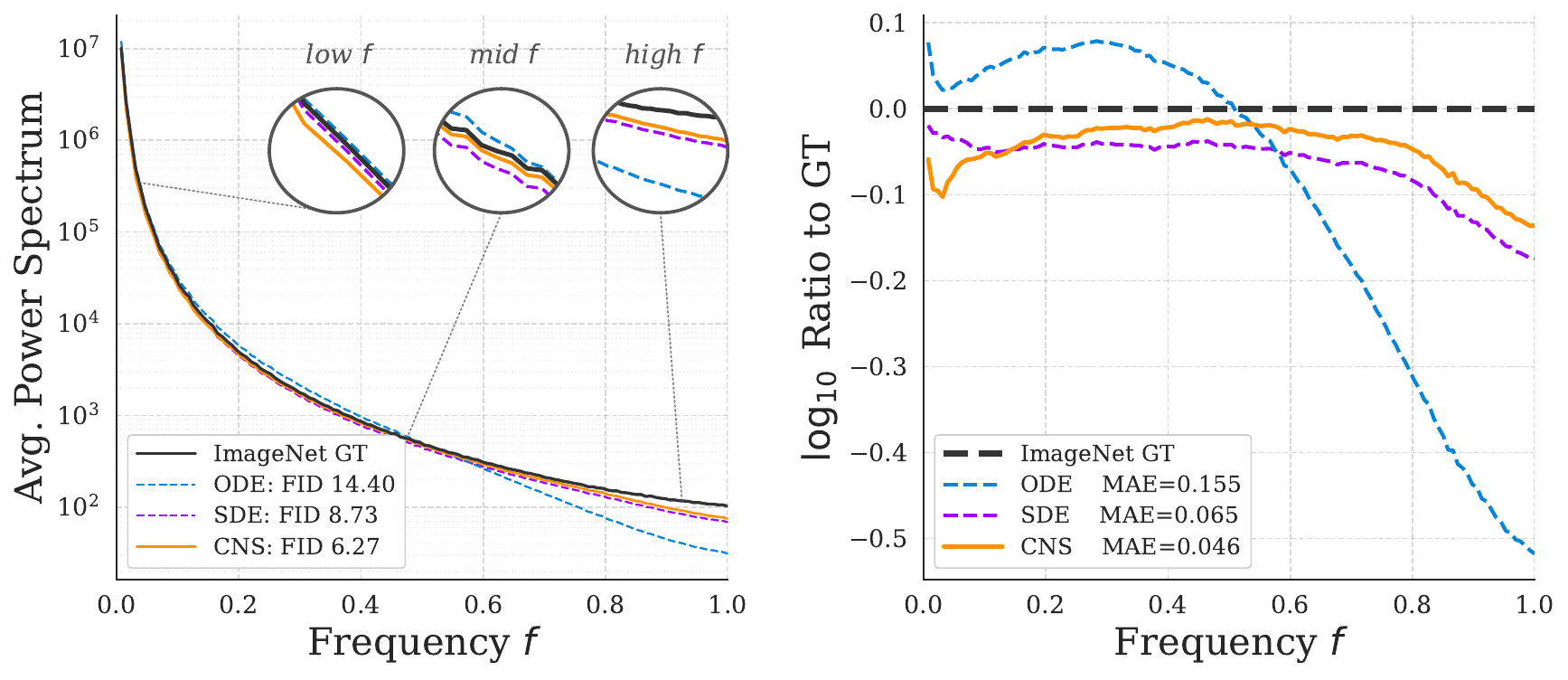}
  \caption{\textbf{The Spectral Gap Across Sampling Methods.} \textit{(Left)} PSDs of the generated distributions versus the PSD of the ground truth ImageNet. Standard ODE sampling over-generates low-frequency structures and under-generates high-frequency details, while standard SDE sampling exhibits an energy deficit across the entire spectrum. \textit{(Right)} The signed $\log_{10}$ error relative to the ground truth (black dashed line). By dynamically reallocating the injected noise budget, CNS better aligns the generated spectrum with the true data manifold, mitigating the spectral gap and achieving the lowest log-space Mean Absolute Error (MAE) across frequencies.}
  \label{fig:spectral_gap}
\end{figure}

As established in recent works \citep{falck2025fourier, adamkiewicz2026pretty, zhang2025enhancing}, a distinct discrepancy exists between the PSD of generated images and the true data manifold---known as the \textit{spectral gap}. Let $S_\text{data}(f)$ and $S_0(f)$ denote the PSDs of the real and generated distributions, respectively. As illustrated in Fig.~\ref{fig:spectral_gap}, neither deterministic ($S^\text{ODE}_0$) nor stochastic ($S^\text{SDE}_0$) sampling perfectly recovers the ground truth $S_\text{data}(f)$. Crucially, resolving this gap requires more than simple post-hoc spectral matching; to achieve true distributional matching, the restored energy must align with the coherent spatial structures of the target manifold.

\begin{figure}[t]
    \centering
    \begin{subfigure}{0.49\textwidth}
        \centering
        \includegraphics[width=0.92\linewidth]{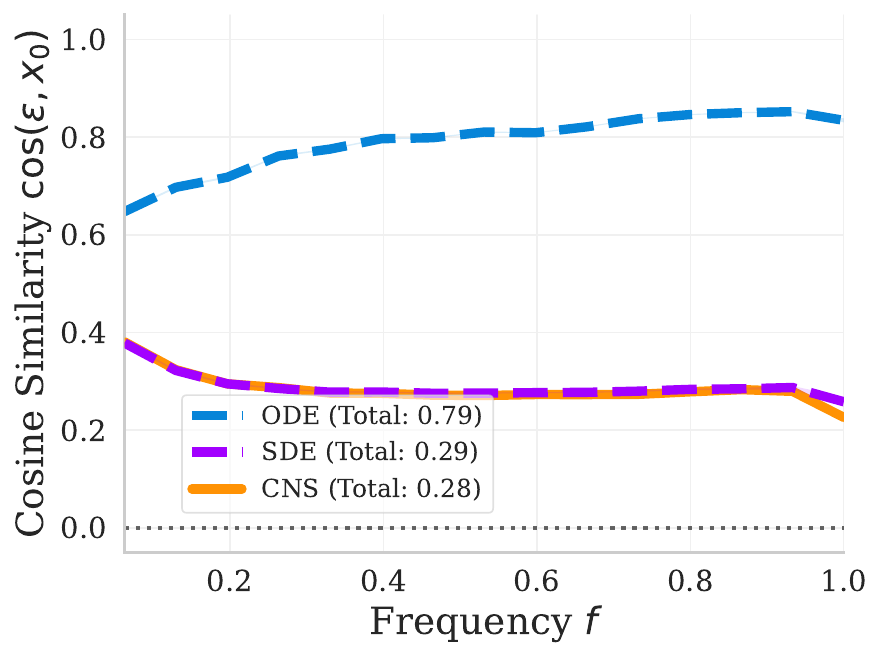}
        \phantomcaption
        \label{fig:cos_sim_ODE_SDE_freq_Dependent}
    \end{subfigure}\hfill
    \begin{subfigure}{0.49\textwidth}
        \centering
        \includegraphics[width=0.92\linewidth]{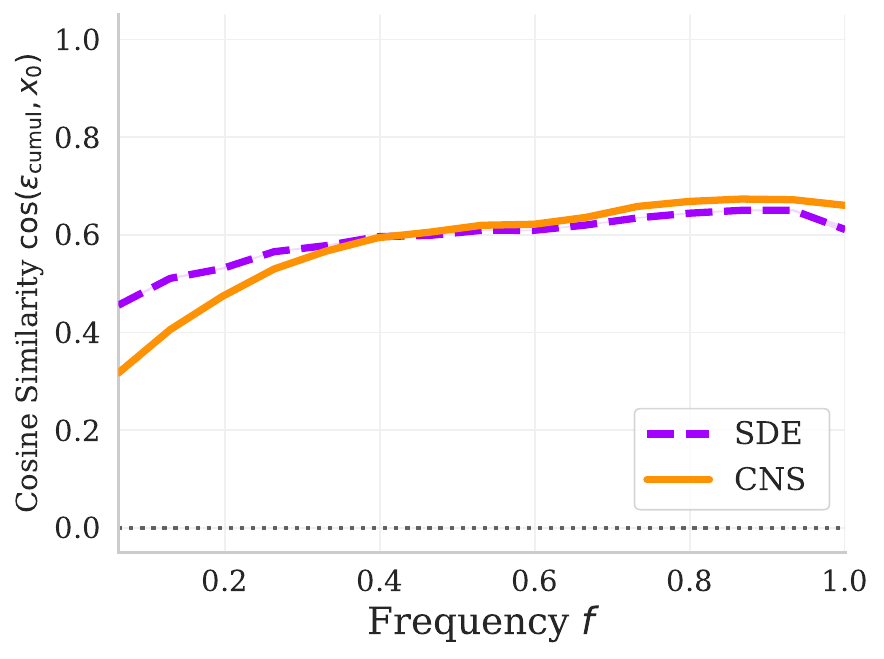}
        \phantomcaption
        \label{fig:cos_sim_SDE_timestep_freq_injected_noise}
    \end{subfigure}
    
    \vspace{-0.2cm}
    \caption{\textbf{Noise Signal Preservation and Transfer.} \textit{(Left)} Initial Noise Persistence. Cosine similarity between initial noise and the final generated image. ODEs strongly preserve structural information across the spectrum; stochastic methods (SDE, CNS) still retain a significant, though reduced, amount of this initial signal. \textit{(Right)} Cumulative Injection Transfer. Cosine similarity between total injected noise ($\epsilon_{\text{cumul}}$) and the final generated image. This shows injected noise structure actively shapes the final features rather than serving as a temporary perturbation. Notably, CNS selectively routes this signal into higher frequency bands.}
    \label{fig:energy_preserved_noise_to_image}
    \vspace{-0.4cm}
\end{figure}

Fortunately, as established in Sec.~\ref{sec:energy_transfer}, diffusion inference fundamentally operates as a partial energy-preserving signal transfer. Beyond preserving the initial noise realization, we found that the diffusion process also maps the stochastic increments injected by SDE solvers directly into corresponding spatial frequencies of the final generated structure. We formalize this non-destructive, frequency-coupled mapping by isolating spatial frequency bands via a Fourier-space band-pass projection operator, $P_b[\cdot]$. This allows us to quantify the structural alignment between the accumulated injected noise, $\epsilon_{\text{cumul}} = \sum_{i=0}^{T-1} g(t_i) d\bar{\mathrm{w}}_{t_i}$, and the final generated image $x_0$. By calculating their expected cosine similarity in Fig.~\ref{fig:cos_sim_SDE_timestep_freq_injected_noise}, we observe a significant positive correlation:
\begin{equation}
    \mathbb{E}\left[ \cos\Big(P_b[\epsilon_{\text{cumul}}], P_b[x_0]\Big) \right] \gg 0
\end{equation}
This strong alignment reveals a powerful theoretical pathway: strategically shaping the spectral profile of the injected noise provides a direct mechanism to steer the PSD of the generated distribution toward the target data manifold.

\noindent \textbf{The Impact of Stochasticity.} \quad To restructure these dynamics, we quantify spectral deviation using the signed log error: $\varepsilon(f) = \log_{10} (S_0(f)/S_{\text{data}}(f))$. As shown in Fig.~\ref{fig:spectral_gap}, comparing deterministic and stochastic solvers reveals that continuous noise injection fundamentally alters the final energy distribution. 
We suggest that this spectral divergence arises from inherent imperfections in the learned score function. During standard Langevin dynamics, the injected noise is not perfectly counterbalanced by the denoising drift. Consequently, score approximation errors cause unintended energy accumulations or deficits over the trajectory (App.~\ref{app:subsec:spectral_gap_origins}).

Crucially, the total stochastic energy injected over the generative trajectory is strictly bounded and mathematically independent of the time discretization (App.~\ref{app:subsec:invariance}). Because we cannot simply scale up the global noise injection to offset deficits without violating the underlying SDE (App.~\ref{app:subsec:variance_conservation}), the stochastic noise acts as a strictly fixed \textit{injected energy budget}: 
\begin{equation}
    \mathcal{E} = \int g^2(t)dt < \infty
\end{equation}
Standard SDEs distribute this budget naively: uniform white noise allocates energy equally across the entire frequency spectrum ($S(f)=1$). By transitioning to targeted \textit{colored noise}, we treat this as a zero-sum game: we dynamically decrease energy allocation for structurally resolved frequency bands, freeing up the budget to inject energy into lagging frequencies. This principled reallocation steers the generated output toward the true data manifold without pushing the intermediate latents out-of-distribution (App.~\ref{app:state_in_distribution}).

\subsection{Colored Noise Sampling (CNS)}
\label{sec:method_CNS}
CNS actively mitigates the spectral gap by repurposing the SDE's stochastic energy leak to steer the generated profile.
As derived in App.~\ref{app:subsec:spectral_gap_origins}, the \textit{effective} energy a generated sample absorbs from noise injection is highly state-dependent. Specifically, the band-wise energy absorption rate depends strictly on the correlation between the current spectral state and the local score error. Because this absorption efficiency varies, uniform white-noise injection is highly suboptimal---it allocates the finite energy budget on frequency modes that are already sufficiently resolved. An optimal strategy must therefore dynamically adapt the noise spectrum to the timestep $t$ and frequency band $f$.

To formalize this active reallocation, we introduce a frequency-dependent scaling weight $\beta_f(t)$ to the standard SDE noise increment. This colored-noise modification scales the stochastic It\^o energy term for a given frequency $f$ from $\frac{1}{2}g^2(t)$ to $\frac{1}{2}g^2(t)\beta_f^2(t)$ (App.~\ref{app:subsec:cns_spectral_impact}). To maintain the overall stability of the generative process, we enforce a strict global variance-conservation constraint, ensuring the average injected energy across all dimensions remains constant: $\frac{1}{D}\sum_{f=1}^{D} \beta_f^2(t) = 1$.
In App.~\ref{app:subsec:cns_spectral_impact}, we demonstrate that a frequency band's capacity to absorb injected stochastic noise into a permanent structure is strictly governed by its progression ratio, tracked by the $\gamma(f,t)$-matrix (Sec.~\ref{sec:spectral_bias}). As a band approaches a fully resolved state ($\gamma(f,t) \to 1$), the score error correlation decays. The network treats excess injected variance primarily as transient energy to be dissipated, severely diminishing the rate of permanent energy conversion.

\definecolor{codeblue}{rgb}{0.25,0.5,0.5}
\definecolor{codesign}{RGB}{0,0,255}
\definecolor{codefunc}{rgb}{0.85,0.18,0.50}
\definecolor{lightpurple}{RGB}{132, 103, 215}
\definecolor{yellowncs}{rgb}{0.87, 0.68, 0.32}

\lstdefinelanguage{PythonFuncColor}{
  language=Python,
  keywordstyle=\color{blue}\bfseries,
  commentstyle=\color{codeblue},
  stringstyle=\color{orange},
  showstringspaces=false,
  basicstyle=\ttfamily\small,
  literate=
    {randn_like}{{\color{codefunc}randn\_like}}{1}
    {_like}{{\color{codefunc}\_like}}{1}
    {bin_radially}{{\color{codefunc}bin\_radially}}{1}
    {ifft2}{{\color{codefunc}ifft2}}{1}
    {linspace}{{\color{codefunc}linspace}}{1}
    {diffusion}{{\color{yellowncs}diffusion}}{1}
    {enumerate}{{\color{codefunc}enumerate}}{1}
    {zeros}{{\color{codefunc}zeros}}{1}
    {randn}{{\color{codefunc}randn}}{1}
    {fft2}{{\color{codefunc}fft2}}{1}
    {real}{{\color{codefunc}real}}{1}
    {std}{{\color{codefunc}std}}{1}
    {sqrt}{{\color{codefunc}sqrt}}{1}
    {model}{{\color{yellowncs}model}}{1}
    {abs}{{\color{codefunc}abs}}{1}
    {*}{{\color{codesign}*}}{1}
    {-}{{\color{codesign}-}}{1}
    {+}{{\color{codesign}+}}{1}
    {torch}{{\color{lightpurple}torch}}{1}
}

\lstset{
  language=PythonFuncColor,
  backgroundcolor=\color{white},
  basicstyle=\fontsize{9pt}{9.9pt}\ttfamily\selectfont,
  columns=fullflexible,
  breaklines=true,
  captionpos=b,
}

\begin{wrapfigure}{r}{0.45\linewidth}
\vspace{-2em}
\centering
\begin{minipage}{0.95\linewidth}

\begin{algorithm}[H]
\caption{CNS Sampling}
\label{alg:cns_sample}
\begin{lstlisting}
# gamma: [T, F] completion matrix
# freq_bins: (H,W) radial bin map
x = torch.randn(x_shape)
for i, t in enumerate(torch.linspace(0, 1, T)):
    w = torch.randn_like(x)
    scale = torch.sqrt(1-gamma[i])
        # colored PSD
    W = torch.fft2(w)*scale[freq_bins]
    w_c = torch.real(ifft2(W))
    w_c = w_c / torch.std(w_c)
    v = model(x, t)
    s = (t*v-x)/(1-t)    # score
    D = diffusion(x, t)
    x += (v+D*s)*dt + torch.sqrt(2*D)*w_c*torch.sqrt(dt)
\end{lstlisting}
\end{algorithm}

\end{minipage}
\vspace{-4em}
\end{wrapfigure}
To maximize the utility of the finite energy budget, CNS dynamically routes energy away from these resolved bands and actively channels it into lagging frequencies with a high structural deficit. We formulate the CNS allocation schedule such that the variance multiplier is strictly proportional to this structural deficit (further details in App.~\ref{app:colored_noise_generated_spectrum}). To satisfy the global energy constraint, we normalize this profile by its Root Mean Square (RMS) across frequencies:
\begin{equation}
    \beta(f,t) = \frac{\sqrt{1-\gamma(f,t)}}{\sqrt{\frac{1}{D}\sum_{f'} \left(1-\gamma(f',t)\right)}}
\end{equation}
This dynamic coloring systematically increases the efficiency of the injected energy, yielding a generated distribution spectrally closer to the true data manifold (Fig.~\ref{fig:spectral_gap}). The exact integration of this schedule into standard SDE solvers is detailed in Alg.~\ref{alg:cns_sample}.
\begin{table}[t]
  \caption{\textbf{Evaluation of Unguided Image Generation.} ImageNet-256 evaluation metrics without Classifier-Free Guidance across different sampling methods.}
  \label{tab:unguided-evaluation}
  \centering
  \small
  \begin{tabular}{l l c c c c c}
    \toprule
    \textbf{Model} & \textbf{Sampler} & \textbf{FID} $\downarrow$ & \textbf{sFID} $\downarrow$ & \textbf{IS} $\uparrow$ & \textbf{Prec.} $\uparrow$ & \textbf{Rec.} $\uparrow$ \\
    \midrule
    \multirow{3}{*}{SiT-XL/2} 
    & ODE & 14.39 & 10.54 & 99.32 & 0.59 & \textbf{0.67} \\
    & SDE & 8.26 & 6.32 & 131.65 & 0.68 & \textbf{0.67} \\
    & CNS (Ours) & \textbf{6.27} & \textbf{4.73} & \textbf{147.33} & \textbf{0.71} & 0.65 \\
    \midrule
    \multirow{3}{*}{JiT-H/16} 
    & ODE & 12.41 & 9.35 & 43.61 & 0.63 & 0.62 \\
    & SDE & 11.88 & 8.64 & 44.30 & 0.64 & 0.63 \\
    & CNS (Ours) & \textbf{8.31} & \textbf{7.48} & \textbf{45.97} & \textbf{0.66} & \textbf{0.65} \\
    \midrule
    \multirow{3}{*}{JiT-B/16} 
    & ODE & 32.39 & 11.81 & 26.60 & 0.47 & 0.61 \\
    & SDE & 36.24 & 14.32 & 25.86 & 0.46 & \textbf{0.63} \\
    & CNS (Ours) & \textbf{26.69} & \textbf{9.67} & \textbf{27.95} & \textbf{0.51} & \textbf{0.63} \\
    \bottomrule
    \vspace{-0.5cm}
  \end{tabular}
\end{table}

\section{Experiments}   
\label{sec:experiments}
We evaluate the performance and robustness of CNS across three key areas: (1) Class-Conditional Generation (Sec.~\ref{subsec:experiments_C2I}), demonstrating superiority over ODE/SDE baselines across pixel and latent spaces, varying solver orders, and CFG settings; (2) Text-to-Image Generation (Sec.~\ref{subsec:experiments_T2I}), integrating CNS into state-of-the-art flow-matching architectures to enhance visual fidelity while preserving strict semantic alignment; and (3) Ablations and Orthogonality, validating our schedule design choices and proving CNS provides orthogonal benefits to models trained with alternative noise distributions. 

\begin{table}[t]
  \caption{\textbf{Evaluation of SiT-XL/2 by Solver Order.} ImageNet-256 evaluation metrics comparing different solvers and sampling methods. Solvers are categorized by their weak convergence order for SDEs (or deterministic order for ODEs). }
  \label{tab:sit-solver-order-evaluation-table}
  \centering
  \small
  \begin{tabular}{l l c c c c c}
    \toprule
    \textbf{Solver} & \textbf{Method} & \textbf{FID} $\downarrow$ & \textbf{sFID} $\downarrow$ & \textbf{IS} $\uparrow$ & \textbf{Precision} $\uparrow$ & \textbf{Recall} $\uparrow$ \\
    \midrule
    \multirow{3}{*}{\shortstack[l]{Euler-Maruyama \cite{maruyama1955continuous}\\ (1st-Order Weak)}} 
    & ODE & 14.39 & 10.54 & 99.32 & 0.59 & \textbf{0.67} \\
    & SDE & 8.26 & 6.32 & 131.65 & 0.68 & \textbf{0.67} \\
    & CNS (Ours) & \textbf{6.27} & \textbf{4.73} & \textbf{147.33} & \textbf{0.71} & 0.65 \\
    \midrule
    \multirow{3}{*}{\shortstack[l]{Heun \cite{heun1900neue, karras2022elucidating}\\ (2nd-Order Weak)}} 
    & ODE & 9.35 & 6.38 & 126.06 & 0.67 & \textbf{0.68} \\
    & SDE & 8.00 & 5.49 & 132.72 & 0.69 & 0.67 \\
    & CNS (Ours) & \textbf{5.99} & \textbf{4.78} & \textbf{149.78} & \textbf{0.71} & 0.65 \\
    \midrule
    \multirow{4}{*}{\shortstack[l]{Stochastic RK \cite{rossler2009second, rossler2010runge}\\ (2nd-Order Weak)}} 
    & SDE (SRK2) & 8.14 & 5.69 & 132.53 & 0.69 & \textbf{0.67} \\
    & SDE (SRK2S) & 8.77 & 6.36 & 129.68 & 0.68 & \textbf{0.67} \\
    & CNS (Ours, SRK2) & \textbf{5.91} & 4.77 & \textbf{149.41} & \textbf{0.71} & 0.65 \\
    & CNS (Ours, SRK2S) & 5.97 & \textbf{4.73} & 148.55 & 0.70 & 0.66 \\
    \midrule
    \multirow{2}{*}{\shortstack[l]{Deterministic RK \cite{hairer2008solving}\\ (5th-Order)}} 
    & \multirow{2}{*}{ODE (dopri5)} & \multirow{2}{*}{9.04} & \multirow{2}{*}{6.04} & \multirow{2}{*}{126.49} & \multirow{2}{*}{0.67} & \multirow{2}{*}{0.67} \\
    & & & & & & \\
    \bottomrule
  \end{tabular}
\end{table}
\begin{table}[t]
\vspace{-0.1cm}
  \caption{\textbf{Evaluation of Guided Sampling (CFG)}  across SiT and JiT architectures on ImageNet-256. }
  \label{tab:merged-cfg-evaluation}
  \centering
  \small
  \begin{tabular}{l l c c c c c c}
    \toprule
    \textbf{Model} & \textbf{Sampler} & \textbf{CFG Scale} & \textbf{FID} $\downarrow$ & \textbf{sFID} $\downarrow$ & \textbf{IS} $\uparrow$ & \textbf{Prec.} $\uparrow$ & \textbf{Rec.} $\uparrow$ \\
    \midrule
    \multirow{3}{*}{SiT-XL/2} & ODE & 1.5 & 2.15 & 4.60 & 258.09 & 0.81 & \textbf{0.60} \\
    & SDE & 1.5 & 2.06 & 4.49 & \textbf{277.50} & \textbf{0.83} & 0.59 \\
    & CNS (Ours) & 1.5 & \textbf{1.98} &\textbf{4.46} & 257.68 & 0.81 & \textbf{0.60} \\
    \midrule
    \multirow{4}{*}{JiT-H/16} & ODE & 2.2 & 3.92 & 7.72 & 62.86 & 0.76 & 0.59 \\
    & SDE & 2.2 & 2.08 & 7.59 & 65.88 & \textbf{0.79} & 0.61 \\
    & CNS (Ours) & 2.2 & \textbf{2.03} & \textbf{6.98} & \textbf{65.89} & \textbf{0.79} & \textbf{0.62} \\
    \midrule
    \multirow{4}{*}{JiT-B/16} & ODE & 3.0 & 5.83 & \textbf{8.71} & 61.47 & 0.81 & 0.43 \\
    & SDE & 3.0 & 4.54 & 11.30 & 63.02 & 0.82 & 0.45 \\
    & CNS (Ours) & 3.0 & {4.39} & 9.48 & \textbf{63.22} & \textbf{0.83} & 0.45 \\
    & CNS (Ours) & 2.6 & \textbf{4.19} & 9.01 & 60.27 & 0.81 & \textbf{0.47} \\
    \bottomrule
  \end{tabular}
  \vspace{-0.5cm}
\end{table}

\subsection{Class-to-Image Generation}
\label{subsec:experiments_C2I}

We evaluate CNS on class-conditional image generation (ImageNet-256 \citep{russakovsky2015imagenet}), comparing it to standard ODE and SDE baselines \citep{song2020score, song2020denoising}. We assess both pixel-space $x$-prediction (JiT-H/16, JiT-B/16 \citep{li2025back}) and latent-space $v$-prediction (SiT-XL/2 \citep{ma2024sit}) models using their official pre-trained weights (Tab.~\ref{tab:unguided-evaluation}). 
We measure generation quality, diversity, and manifold coverage using established metrics: Fréchet Inception Distance (FID) \citep{heusel2017gans}, spatial FID (sFID) \citep{nash2021generating} for high-frequency structural assessment, Inception Score (IS) \citep{salimans2016improved}, and Precision and Recall \citep{kynkaanniemi2019improved}.
For fair comparison against reported baselines, we utilize originally published metric values wherever available. To maintain identical evaluation conditions across reproduced methods, we fix the initial seed across all methods. We match the sampling steps of the original works: 250 for SiT-XL/2 and 50 for JiT models. All quantitative metrics are computed using 50,000 generated samples against the standard 10,000 reference images. Qualitative visual comparisons of CNS against SDE and ODE baselines are provided in App.~\ref{sec:appendix_Visual_compare}. 

\begin{wrapfigure}{r}{0.4\textwidth}
  \vspace{-0em} 
  \centering
  \includegraphics[width=0.99\linewidth]{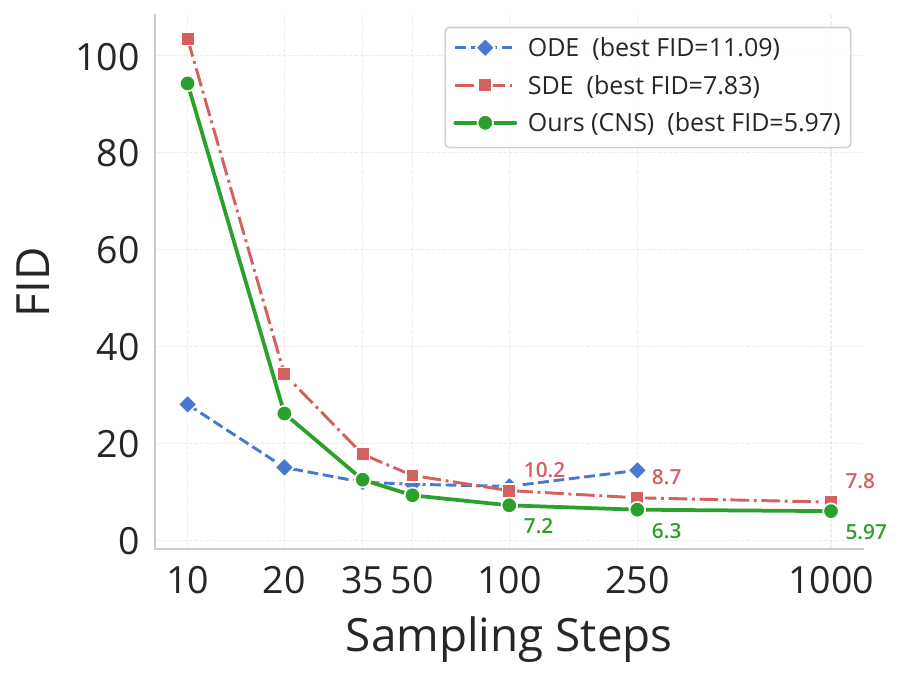}
  \caption{FID-50K vs sampling steps for different samplers.}
  \label{fig:SiT_sampling_steps_results}
  \vspace{-1em} 
\end{wrapfigure}

Furthermore, Fig.~\ref{fig:SiT_sampling_steps_results} evaluates our sampler across varying time discretizations. Once a sufficient number of steps for proper stochastic simulation is reached, CNS consistently outperforms standard sampling methods. Further details are provided in App.~\ref{app:discretization_details}.

\noindent \textbf{Solver Order.} \quad While deterministic ODEs rely on standard Taylor expansions, SDE solvers require It\^{o}-Taylor expansions and are categorized by \textit{strong} (pathwise) and \textit{weak} (distributional) convergence orders. Because generative metrics evaluate distributional alignment, weak convergence is our primary focus. As shown in Tab.~\ref{tab:sit-solver-order-evaluation-table}, we evaluate CNS across solvers with varying weak orders (1st-order Euler-Maruyama; 2nd-order Heun, SRK2, SRK2S). CNS outperforms all tested solvers on SiT-XL/2. Mathematical overviews of these solvers are in App.~\ref{app:solver_details}.

\noindent \textbf{Classifier-Free Guidance (CFG).} \quad CFG \citep{ho2022classifier} significantly improves sample fidelity by extrapolating the conditional prediction away from the unconditional baseline. In Tab.~\ref{tab:merged-cfg-evaluation}, we demonstrate that CNS consistently outperforms standard ODE and SDE samplers under CFG across SiT-XL/2, JiT-H/16, and JiT-B/16 (using Euler, 250 steps for SiT, 50 for JiT). When optimal CFG scales diverge between methods, we report the best-performing scale for both the baseline and CNS. Qualitative comparisons of CFG-enhanced generation are in Fig.~\ref{fig:teaser}.

 \noindent \textbf{Orthogonality to Alternative Noise Training.} \quad Recent methods \citep{scimeca2025learning, falck2025fourier, huang2024blue} alter the training framework to explicitly exploit spectral bias. To confirm CNS is not rendered redundant by such modifications, we integrate our sampler into Blue Noise for Diffusion Models (BNDM) \citep{huang2024blue}, which trains a U-Net \citep{ronneberger2015u} with a time-evolving white-to-blue noise distribution. As shown in Tab.~\ref{tab:bndm-all_datasets-evaluation-table}, applying CNS to BNDM's pre-trained Iterative $\alpha$-(de)Blending (IADB) \citep{heitz2023iterative} models still significantly improves generation quality. Evaluated across two $64^2$ datasets ($30,000$ samples), this confirms CNS provides orthogonal inference-time benefits even to models inherently designed to leverage spectral bias. Further details are in App.~\ref{app:bndm_details}.

\begin{table}[t]
  \caption{\textbf{FID-30K ($\downarrow$) on $64^2$ Datasets.} Comparison of baseline models and BNDM variants. }
  \label{tab:bndm-all_datasets-evaluation-table}
  \centering
  \small
  \begin{tabular}{l c c c c c c c}
    \toprule
    & \multicolumn{4}{c}{\textbf{Prior Methods}} & \multicolumn{3}{c}{\textbf{BNDM Framework}} \\
    \cmidrule(lr){2-5} \cmidrule(lr){6-8}
    \textbf{Dataset} & \textbf{IHDM} & \textbf{DDPM} & \textbf{DDIM} & \textbf{IADB} & \textbf{ODE} & \textbf{SDE} & \textbf{CNS (Ours)} \\
    \midrule
    AFHQ Cat & 11.02 & 9.75 & 9.82 & 9.19 & 7.95 & 18.80 & \textbf{7.49} \\
    LSUN Church & 17.76 & 13.07 & 16.46 & 13.12 & 10.16 & 66.71 & \textbf{8.70} \\
    \bottomrule
  \end{tabular}
\end{table}

\subsection{Text-to-Image Generation}

\label{subsec:experiments_T2I}
Moving beyond class-conditional synthesis, we demonstrate the broad applicability of our method by seamlessly integrating it into complex downstream generation tasks. Specifically, we apply CNS as a plug-and-play sampler substitution within state-of-the-art Text-to-Image (T2I) flow-matching architectures: FLUX.1-dev \citep{flux2024} and FLUX.2-klein \citep{flux-2-2025}.

\begin{wraptable}{r}{0.52\textwidth}
  \centering
  \caption{\textbf{Quantitative evaluation on DrawBench.} Evaluated at 50 steps, CFG scales $w=3.5,~4.0$ for FLUX1, FLUX2, respectively.}
  \label{tab:drawbench_flux_results}
  \small
  \setlength{\tabcolsep}{4pt} 
  \begin{tabular}{l@{} c@{~} c@{~} c@{}} 
    \toprule
    \textbf{Sampler} & \textbf{ImageReward} $\uparrow$ & \textbf{CLIPScore} $\uparrow$ & \textbf{Aesthetic} $\uparrow$ \\
    \midrule
    \multicolumn{4}{c}{\textit{FLUX.1-dev}} \\
    \midrule
    ODE & 0.965 & 0.681 & 5.787 \\
    SDE & 0.990 & 0.689 & 5.804 \\
    CNS (Ours) & \textbf{1.012} & \textbf{0.693} & \textbf{5.812} \\ 
    \midrule
    \multicolumn{4}{c}{\textit{FLUX2.klein}} \\
    \midrule
    ODE & 0.984 & \textbf{0.735} & 5.233 \\
    SDE & 0.924 & 0.733 & 5.291 \\
    CNS (Ours) & \textbf{1.005} & \textbf{0.735} & \textbf{5.295} \\
    \bottomrule
  \end{tabular}
  \vspace{-1em} 
\end{wraptable}

We evaluate this integration by comparing CNS against both standard white-noise SDEs and the deterministic ODEs that serve as the default solvers for these models. We evaluate generation quality and text-alignment across two comprehensive prompt benchmarks. DrawBench \citep{saharia2022photorealistic} probes specific T2I failure modes like complex text rendering (Tab.~\ref{tab:drawbench_flux_results}), while GenEval \citep{ghosh2023geneval} tests precise compositional attributes such as object counts and spatial positioning (Tab.~\ref{tab:geneval_flux_results}). Performance is measured using ImageReward \citep{xu2023imagereward} for human preference, CLIPScore \citep{hessel2021clipscore} for semantic consistency, and Aesthetic Score \citep{schuhmann2022improved} for visual appeal. 
Crucially, these evaluations confirm that the dynamic stochastic modifications introduced by CNS enhance overall visual fidelity without degrading the underlying model's text comprehension or corrupting complex compositional instructions.

\subsection{Ablation Study}
\label{subsec:ablation_study}
\begin{wraptable}{r}{0.43\textwidth}
  \centering
  \vspace{-1em}
  \caption{\textbf{Key Ablation Studies (FID-10K).} 
  Full ablation results are in Appendix \ref{app:subsec:ablation_study_full}. }
  \label{tab:ablation_studies_main}
  \small
  \begin{tabular}{l@{} c@{} c@{} c@{}}
    \toprule
    \textbf{Scenario} & \textbf{FID} $\downarrow$~ & ~\textbf{sFID} $\downarrow$ & \textbf{IS} $\uparrow$ \\
    \midrule
    \textbf{CNS (Ours)} & \textbf{9.61} & \textbf{18.17} & \textbf{143.20} \\
    \midrule
    White-noise SDE & 11.82 & 19.15 & 107.75 \\
    Deterministic ODE & 17.05 & 23.57 & 98.43 \\
    \midrule
    50\% White Noise & 10.64 & 19.08 & 136.36 \\
    Shuffled Schedule & 10.46 & 19.03 & 138.02 \\
    Constant Spectrum 
    & 10.53 & 19.14 & 137.63 \\
    Scale 0.90 Energy & 16.17 & 42.03 & 111.29 \\
    Scale 1.05 Energy & 20.46 & 29.73 & 96.17 \\
    mBm 
    & 11.88 & 19.62 & 130.22 \\
    \bottomrule
    \vspace{-1.5em}
  \end{tabular}
\end{wraptable}
We conduct an ablation study on SiT-XL/2 (Euler, 250 steps), summarized in Tab.~\ref{tab:ablation_studies_main}. While several alternative configurations outperform the standard white-noise SDE, our derived CNS formulation consistently achieves the highest overall fidelity. Specifically, we demonstrate that scaling the injected variance without adhering to our normalization constraint significantly degrades quality, empirically validating the necessity of a strictly finite energy budget. Furthermore, perturbing the CNS schedule, via partial white-noise corruption, static colored spectra, or temporal shuffling, consistently yields inferior results, confirming the optimality of our dynamic, state-dependent allocation. Finally, we establish that replacing our schedule with Multifractional Brownian Motion (mBm) \citep{peltier1995multifractional}, a process generating time-varying colored noise via a shifting Hurst parameter $H(t)$, also falls short of CNS performance. Further mathematical details for all ablations are provided in App.~\ref{app:subsec:ablation_study_full}. 
\section{Conclusion}
\label{sec:conclusion}
In this work, we address a fundamental inefficiency in standard diffusion SDE solvers: the uniform injection of white noise, which ignores the model's inherent spectral bias and squanders the finite generative energy budget. By reconceptualizing SDE inference as a targeted energy transfer, we introduce Colored Noise Sampling (CNS), a novel stochastic sampler. CNS actively exploits spectral bias by dynamically reallocating injected noise toward structurally unresolved frequency bands. As a strictly plug-and-play sampler substitution at inference time, CNS systematically steers the generated distribution toward the true data manifold. Extensive experiments validate that CNS dramatically outperforms standard baselines across diverse architectures, yielding massive FID reductions and enhanced visual fidelity without requiring any model retraining.
\paragraph{Limitations and Future Work}
\label{app:limitations_future_work}
The primary limitation of CNS is its reliance on an SDE framework, rendering it incompatible with deterministic ODE solvers. Because stochastic sampling intrinsically requires a high step budget to prevent discretization error accumulation, standard ODEs remain preferable for ultra-fast inference. Future work will explore extending frequency-dependent energy routing into deterministic paradigms for low-step sampling, and applying CNS to video generation to leverage the temporal frequency dimension.

\clearpage

\bibliography{references}
\bibliographystyle{abbrvnat}

\clearpage
\appendix

\section{Theoretical Constraints on Stochastic Energy Injection}
\label{app:energy_constraints}
In the main text, we establish that the stochastic noise injected during the generative process acts as a strictly fixed \textit{injected energy budget}. This appendix provides the formal derivations underpinning this core premise. First, in App.~\ref{app:subsec:invariance}, we demonstrate that the total injected stochastic energy is mathematically finite and strictly invariant to the time discretization of the solver. Building upon this, in App.~\ref{app:subsec:variance_conservation}, we show that this finite energy budget cannot be globally scaled to offset spectral deficits without violating the underlying SDE and breaking the deterministic convergence to the target data manifold. Together, these theoretical constraints necessitate our zero-sum, frequency-band redistribution approach.

\subsection{Invariance of Total Injected Energy to Timestep Discretization}
\label{app:subsec:invariance}
In this section, we demonstrate that for a given continuous diffusion schedule $g(t)$ over a generative interval $[t_0, t_1]$, the total stochastic energy injected by the noise process is finite and strictly invariant to the number of discretization steps used to integrate the reverse SDE. This formalizes our conceptualization of the SDE's stochasticity as a strictly fixed, global injected energy budget.

Consider the continuous reverse-time SDE \citep{anderson1982reverse, song2020score}:
\begin{equation}
dx_t = \mu(x_t, t)dt + g(t)d\bar{\mathrm{w}}
\end{equation}
where the deterministic drift is defined as $\mu(x_t, t) = \left[v(x_t,t) - \frac{1}{2}g^2(t)\nabla_{x_t} \log p_t(x_t)\right]$, $d\bar{\mathrm{w}}$ is the standard reverse-time Wiener process, and $g: [t_0, t_1] \to \mathbb{R}_{\ge 0}$ is a continuous and bounded diffusion coefficient.

Let our discrete sampler be evaluated over $N$ uniform integration steps using standard Euler-Maruyama numerical integration \citep{kloeden2011numerical}:
\begin{equation}
x_{k+1} = x_k + \mu(x_k,t_k)\Delta t + g(t_k)\Delta \bar{\mathrm{w}}_k, \qquad \Delta \bar{\mathrm{w}}_k \sim \mathcal{N}(0, \Delta t \mathbf{I})
\end{equation}
Isolating a single spatial dimension for clarity, we define the injected stochastic increment at step $k$ as:
\begin{equation}
\eta_k := g(t_k)\Delta \bar{\mathrm{w}}_k
\end{equation}
and the total expected injected variance per dimension over the entire trajectory as:
\begin{equation}
\mathcal{E}_N := \sum_{k=0}^{N-1} \mathbb{E}[\eta_k^2]
\end{equation}

\paragraph{Proof of Convergence.}
Because the diffusion coefficient $g(t_k)$ is deterministic at step $k$, the expected variance of the increment is:
\begin{equation}
\mathbb{E}[\eta_k^2] = \mathbb{E}\left[ \left(g(t_k)\Delta \bar{\mathrm{w}}_k\right)^2 \right] = g^2(t_k)\mathbb{E}[\Delta \bar{\mathrm{w}}_k^2] = g^2(t_k)\Delta t
\end{equation}
Therefore, the total expected variance is:
\begin{equation}
\mathcal{E}_N = \sum_{k=0}^{N-1} g^2(t_k)\Delta t
\end{equation}
This formulation is exactly the left Riemann sum of $g^2(t)$ over the interval $[t_0, t_1]$. By the continuity of $g(t)$, we can take the continuous-time limit:
\begin{equation}
\lim_{N\to\infty} \mathcal{E}_N = \int_{t_0}^{t_1} g^2(t)dt =: \mathcal{E}
\end{equation}
Because standard diffusion schedules $g(t)$ are strictly bounded on a finite interval \citep{song2020score}, the integral $\mathcal{E}$ evaluates to a finite constant.

\paragraph{Finite-$N$ Discretization Error.}
Let $\mathrm{Err}(N) := \left|\mathcal{E}_N - \mathcal{E}\right|$ denote the variance discrepancy for a discrete solver. If $g^2(\cdot) \in C^1([t_0, t_1])$, the bounded error of the left Riemann sum guarantees:
\begin{equation}
\mathrm{Err}(N) \le \frac{(t_1-t_0)^2}{2N} \max_{t\in[t_0,t_1]}\left| \frac{d}{dt} g^2(t) \right|
\end{equation}
This demonstrates that the convergence is $\mathcal{O}(1/N)$. Consequently, up to a minor, strictly bounded numerical integration error, any chosen number of steps $N$ draws from the exact same finite pool of total injected energy.

\subsection{Theoretical Demand for Injected Noise Variance Conservation}
\label{app:subsec:variance_conservation}
In this section, we demonstrate that a Langevin dynamics formulation in which the injected global noise variance is misaligned with the true score function---even under the assumption of an exact oracle score---yields a sampling process that fundamentally fails to converge to the target data distribution, $p_{\text{data}}$.

\paragraph{The Modified Reverse SDE and its Equivalent Flow.}
In continuous-time diffusion frameworks, the reverse generative process is governed by the stochastic differential equation (SDE) \citep{song2020score}:
\begin{equation}
dx_t = \left[ v(x_t,t) - \frac{1}{2}g^2(t)\nabla_{x_t} \log p_t(x_t) \right] dt + g(t) d\bar{\mathrm{w}}
\end{equation}
where $v(x_t,t)$ is the deterministic drift, $g(t)$ is the diffusion coefficient, and $d\bar{\mathrm{w}} \sim \mathcal{N}(0, dt \mathbf{I})$ is the reverse-time Wiener process.

Let us introduce a constant global variance multiplier $\beta$ to the stochastic increment. The modified SDE becomes:
\begin{equation}
dx_t = \left[ v(x_t,t) - \frac{1}{2}g^2(t)\nabla_{x_t} \log p_t(x_t) \right] dt + \beta g(t) d\bar{\mathrm{w}}
\end{equation}
To analyze the exact marginal distributions generated by this modified SDE, denoted $\rho_t(x_t)$, we rely on the continuous-time equivalence between SDEs and ODEs. For any reverse-time SDE of the form $dx_t = \mu(x_t, t) dt + \sigma(t) d\bar{\mathrm{w}}$, there exists a deterministic Probability Flow ODE that shares the identical marginal density $\rho_t(x_t)$ at all times \citep{anderson1982reverse}:
\begin{equation}
dx_t = \left[ \mu(x_t, t) + \frac{1}{2}\sigma^2(t) \nabla_{x_t} \log \rho_t(x_t) \right] dt
\end{equation}
Substituting our modified drift and diffusion terms into this relation yields the equivalent PF-ODE for our perturbed system:
\begin{equation}
dx_t = \left[ v(x_t,t) - \frac{1}{2}g^2(t)\nabla_{x_t} \log p_t(x_t) + \frac{1}{2} \beta^2 g^2(t) \nabla_{x_t} \log \rho_t(x_t) \right] dt
\end{equation}
If we enforce the requirement that this modified process successfully reaches the target data manifold exactly along the true marginal trajectory, we must have $\rho_t(x_t) = p_t(x_t)$ for all $t$. Under this condition, the equivalent flow simplifies to:
\begin{equation}
dx_t = \left[ v(x_t,t) - \frac{1}{2}g^2(t) \left(1 - \beta^2\right) \nabla_{x_t} \log p_t(x_t) \right] dt
\end{equation}
However, the true target PF-ODE that corresponds to the unperturbed forward process is known to be:
\begin{equation}
dx_t = v(x_t, t)\, dt
\end{equation}
This identity is recovered directly from the SDE-to-PF-ODE relation applied to the original reverse SDE ($\beta = 1$): the explicit score subtraction $-\tfrac{1}{2}g^2 \nabla_{x_t}\log p_t$ is exactly cancelled by the implicit Fokker--Planck correction $+\tfrac{1}{2}g^2 \nabla_{x_t}\log p_t$ contributed by the diffusion term, leaving only $v(x_t,t)$.

Equating the drift of our modified process with this target requires:
\begin{equation}
\tfrac{1}{2}g^2(t)\bigl(1 - \beta^2\bigr)\, \nabla_{x_t} \log p_t(x_t) = 0
\end{equation}
Since $g(t) > 0$ and the score $\nabla_{x_t}\log p_t$ is generically nonzero along the generative trajectory, this constraint is satisfied strictly if and only if:
\begin{equation}
\beta^2 = 1
\end{equation}
This mathematically confirms that any global rescaling of the injected stochastic noise disrupts the deterministic flow toward $p_{\text{data}}$.

\paragraph{The Rigid Coupling of Transport and Geometry.}
To rigorously justify why a mismatch ($\rho_t \neq p_t$) causes the process to fail, we examine the continuity equation governing the mass transport of the system. Let $v^*(x_t,t)$ denote the exact PF-ODE vector field. The temporal evolution of the mismatched density $\rho_t$ driven by this static vector field is:
\begin{equation}
\begin{split}
\frac{\partial \rho_t}{\partial t} &= -\nabla_{x_t} \cdot \Big( v^*(x_t,t) \rho_t(x_t) \Big) \\
&= -v^*(x_t,t) \cdot \nabla_{x_t} \rho_t(x_t) - \rho_t(x_t) \nabla_{x_t} \cdot v^*(x_t,t)
\end{split}
\end{equation}
This expansion exposes a fundamental structural incompatibility:
\subparagraph{Advection ($v^*\cdot\nabla_{x_t} \rho_t$):} The translation of mass acts upon the spatial gradient of the \textit{mismatched} distribution $\rho_t$.
\subparagraph{Compression/Expansion ($\rho_t \nabla_{x_t} \cdot v^*$):} The local expansion and compression of probability mass is controlled by the score field, which is specifically tailored to the geometry of the true data distribution $p_t$.

Because $\rho_t$ possesses a different spatial gradient than $p_t$, the static local flow routes mass incorrectly. An area in the latent space might receive an influx of probability mass dictated by $v^*$ expecting the geometry of $p_t$, but instead encounters the density of $\rho_t$, inevitably creating artificial bottlenecks or overshoots. The score field does not act as a global attractor; rather, it serves as a rigidly coupled transport mechanism that remains valid only when the state distribution adheres strictly to its corresponding marginal trajectory.

\paragraph{Practical Implications of $\beta^2 \neq 1$.}
\label{app:subsec:using_under_injection}
While $\beta^2 = 1$ is required for perfect distributional matching, the practical failure modes diverge depending on the direction of the scaling error:
\begin{itemize}
    \item \textbf{Over-injection ($\beta^2 > 1$):} The excessive stochastic exploration outpaces the restorative capacity of the score-based drift. The state rapidly diffuses off the true manifold, resulting in severe degradation of sample fidelity.
    \item \textbf{Under-injection ($\beta^2 < 1$):} The restorative gradient steps overpower the diminished stochastic exploration. While this theoretically violates perfect marginal matching, the samples may not immediately drift off-manifold. Instead, the generative distribution aggressively contracts around local high-density modes. This preserves the visual coherence of individual samples but results in mode concentration and a measurable loss of generative diversity.
\end{itemize}
In App.~\ref{app:state_in_distribution} and App.~\ref{app:subsec:cns_spectral_impact} we demonstrate that in diffusion models, anchoring the density $\rho_t$ to the true probability path requires relaxing the strict per-dimension variance constraint. Instead, the model enforces an average variance across all dimensions, satisfying the following condition:
\begin{equation}
\frac{1}{D}\sum_{f=1}^{D} \beta_f^2(t) = 1
\end{equation}
\subsection{Model Robustness to Out-of-Distribution Spectral States}
\label{app:state_in_distribution}

A fundamental assumption in our Colored Noise Sampling (CNS) framework is that actively modulating the per-frequency Signal-to-Noise Ratio (SNR) via the scaling vector $\beta_f(t)$ does not push the intermediate generative states so far Out-Of-Distribution (OOD) that the network fails. In this section, we demonstrate that this robustness is an inherent property of diffusion models. Specifically, because of the network's inductive spectral bias, standard inference trajectories intrinsically operate on highly OOD spectral states, yet the model reliably resolves these states into the true data distribution. 

\paragraph{The Theoretical Training Paradigm.}
During standard continuous-time training (e.g., under a linear Flow Matching schedule, for simplicity), the model is optimized over exact marginals $p_t(x_t)$. Given a clean data sample $x_0 \sim p_{\text{data}}$ and a noise sample $\epsilon \sim \mathcal{N}(0, I)$, the intermediate state flows from pure noise at $t=1$ to clean data at $t=0$ via:
\begin{equation}
x_t = (1-t)x_0 + t\epsilon
\end{equation}
Applying the Fourier transform yields the spectral composition of the training marginals:
\begin{equation}
\widehat{x}_t(f) = (1-t)\widehat{x}_0(f) + t\widehat{\epsilon}(f)
\end{equation}
Later established in App.~\ref{app:subsubsec:state_error_cor_psd_gap}, the expected clean data spectrum exhibits a power-law distribution $|\widehat{x}_0(f)| \approx C/f^{\omega_\text{amp}}$, while the noise prior is spectrally flat $|\widehat{\epsilon}(f)| \approx 1$. 

If the generative process were to perfectly trace this theoretical training path, the proportion of the resolved image signal at any frequency band $f$ would evolve strictly linearly across time. By isolating the data term and normalizing, the expected target progression ratio $\gamma_{\text{target}}(f,t)$ is defined as:
\begin{equation}
\gamma_{\text{target}}(f,t) = \frac{\mathbb{E}[|\widehat{x}_t(f) - t\widehat{\epsilon}(f)|]}{\mathbb{E}[|\widehat{x}_0(f)|]} = 1-t
\end{equation}
Under this idealized paradigm, every frequency band accumulates image information uniformly, maintaining the exact SNR balance dictated by the training marginals.

\paragraph{Inference-Time Spectral Divergence.}
However, during standard inference, the model does not generate the image uniformly. As quantified by the $\gamma$-matrix (Sec.~\ref{sec:spectral_bias}), the neural network exhibits a profound spectral bias. 

Both when initiated from perfect pure noise and integrated with high precision, and when given some intermediate state of the interpolated distribution, the empirical progression of the frequency bands, $\gamma_{\text{actual}}(f,t)$, drastically deviates from the linear target $\gamma_{\text{target}}(f,t) = 1-t$. Low-frequency bands converge much faster than the training paradigm dictates, resolving their structural energy early in the generation process ($t \gg 0$), while high-frequency bands remain noise-dominated until the very end of the trajectory.

\paragraph{Implications for Colored Noise Sampling.}
This empirical divergence reveals a critical operational reality: \textit{during standard inference, the intermediate latents $x_t$ are already severely out-of-distribution with respect to the training marginal's PSD and SNR} \cite{falck2025fourier}.

\definecolor{codeblue}{rgb}{0.25,0.5,0.5}
\definecolor{codesign}{RGB}{0,0,255}
\definecolor{codefunc}{rgb}{0.85,0.18,0.50}
\definecolor{lightpurple}{RGB}{132, 103, 215}
\definecolor{yellowncs}{rgb}{0.87, 0.68, 0.32}

\lstdefinelanguage{PythonFuncColor}{
  language=Python,
  keywordstyle=\color{blue}\bfseries,
  commentstyle=\color{codeblue},
  stringstyle=\color{orange},
  showstringspaces=false,
  basicstyle=\ttfamily\small,
  literate=
    {randn_like}{{\color{codefunc}randn\_like}}{1}
    {bin_radially}{{\color{yellowncs}bin\_radially}}{1}
    {ifft2}{{\color{codefunc}ifft2}}{1}
    {mean}{{\color{codefunc}mean}}{1}
    {cat}{{\color{codefunc}cat}}{1}
    {clamp}{{\color{codefunc}clamp}}{1}
    {linspace}{{\color{codefunc}linspace}}{1}
    {diffusion}{{\color{yellowncs}diffusion}}{1}
    {enumerate}{{\color{codefunc}enumerate}}{1}
    {zeros}{{\color{codefunc}zeros}}{1}
    {randn}{{\color{codefunc}randn}}{1}
    {fft2}{{\color{codefunc}fft2}}{1}
    {real}{{\color{codefunc}real}}{1}
    {std}{{\color{codefunc}std}}{1}
    {sqrt}{{\color{codefunc}sqrt}}{1}
    {model}{{\color{yellowncs}model}}{1}
    {abs}{{\color{codefunc}abs}}{1}
    {*}{{\color{codesign}*}}{1}
    {**}{{\color{codesign}**}}{1}
    {/}{{\color{codesign}/ }}{1}
    {-}{{\color{codesign}-}}{1}
    {+}{{\color{codesign}+}}{1}
    {torch}{{\color{lightpurple}torch}}{1}
}

\lstset{
  language=PythonFuncColor,
  backgroundcolor=\color{white},
  basicstyle=\fontsize{9pt}{9.9pt}\ttfamily\selectfont,
  columns=fullflexible,
  breaklines=true,
  captionpos=b,
}

\begin{wrapfigure}{r}{0.56\linewidth}
\vspace{-2em}
\centering
\begin{minipage}{0.95\linewidth}

\begin{algorithm}[H]
\caption{$\gamma(f,t)$ Matrix Computation}
\label{alg:gamma_mat_calc}
\begin{lstlisting}
# ODE traj x: [T, B, C, H, W],
# x[0]=noise, x[T-1]=image
gamma_sum = torch.zeros(T, F)
for x in ODE_batches:
    t = torch.linspace(0, 1, T)
    v = (x[1:] - x[:-1]) / dt  # [T-1,B,C,H,W]
    xp = x[:-1]+(1-t[:-1])*v   # clean pred [T-1]
    xp = torch.cat([xp, x[-1:]])     # append final [T]
    X  = torch.fft2(xp)
    Xf = torch.fft2(x[-1])     # final spectrum
    g = 1 - torch.abs(X-Xf)**2 / torch.abs(Xf)**2
    g = torch.clamp(g, 0, 1).mean(dim=C)  # ch. avg -> [T,B,H,W]
    gamma_sum += bin_radially(g).mean(dim=B)
gamma = gamma_sum / N
\end{lstlisting}
\end{algorithm}

\end{minipage}
\vspace{-1em}
\end{wrapfigure}
At an intermediate timestep $t$, the theoretical training marginal $p_t(x)$ expects the low frequencies to be only partially built ($\sim 1-t$). In reality, the generated state $x_t^{\text{gen}}$ possesses heavily saturated low frequencies ($\gamma_{\text{actual}} \approx 1$). Despite this massive spectral mismatch—where the local SNR of specific bands completely violates the theoretical training expectation—the network's learned score function $s_\theta(x_t, t)$ does not collapse. It continues to stably process this OOD spectrum, utilizing the combined velocity/score dynamics to reliably push the trajectory toward the true image distribution rather than reverting to noise.

This inherent spectral robustness directly validates the core mechanical assumption of Colored Noise Sampling. Because the model natively handles and corrects intermediate states with perturbed frequency-wise SNR distributions, intentionally shaping the injected stochastic energy via our variance scaling operator $\beta_f(t)$ operates safely within the model's established envelope of robustness. By ensuring that the global variance budget is strictly conserved ($\frac{1}{D}\sum \beta_f^2 = 1$), CNS leverages the network's inherent capacity to convert specifically routed frequency-band energy into coherent image structures without inducing catastrophic OOD failure.

\section{Theoretical Analysis of Spectral Dynamics and Colored Noise}
\label{app:spectral_dynamics_and_cns}
In this appendix, we present a unified theoretical analysis of the spectral dynamics governing generative diffusion models. First, in App.~\ref{app:subsec:spectral_gap_origins}, we investigate the mechanistic origins of the observed spectral gap between ODE and SDE samplers, demonstrating how pathwise energy trajectories diverge due to imperfect score approximation. Building upon this foundation, in App.~\ref{app:subsec:cns_spectral_impact}, we formalize how Colored Noise SDEs could strategically reshape this continuous energy evolution on a strictly band-wise basis. We show that while ideal score conditions enforce a strict zero-sum energy redistribution, the time-dependent decay of state-error correlation allows CNS to fundamentally circumvent this constraint. This enables CNS to permanently and constructively alter the generated spectrum by dynamically routing variance to frequencies where structural conversion efficiency is maximized.

\subsection{Theoretical Origins of the Generated Distributions Spectral Difference}
\label{app:subsec:spectral_gap_origins}
To identify the mechanism behind the observed generated distribution PSDs differences (see Fig.~\ref{fig:spectral_gap}), we theoretically analyze the temporal evolution of frequency-band energy during sampling for both ODE and SDE trajectories. The goal is to characterize \emph{where} (which frequency bands and timesteps) and \emph{how} (rate and direction of energy transfer) the two samplers diverge. This comparison provides a mechanistic explanation for why their final generated spectra differ and, in turn, clarifies the source of their quality gap.

Although the Probability Flow ODE and reverse-time SDE are constructed to share identical marginals $p_t(x)$ at each time $t$, their \emph{pathwise} energy dynamics differ substantially due to the network's imperfect approximation of the score. Let the state energy at time $t$ be defined as:
\begin{equation}
E_t := \frac{1}{2}||x_t||_2^2
\end{equation}
The factor of $\frac{1}{2}$ is a standard normalization that ensures $\nabla_x(\frac{1}{2}\|x\|_2^2) = x$, which cleanly aligns the energy drift expressions.

\subsubsection{Pathwise Energy Dynamics in Continuous-Time Sampling}
Let $v(x_t, t)$ denote the deterministic drift of the PF-ODE. The ODE trajectory is $dx_t = v(x_t,t)\,dt$. Applying standard differentiation, the expected energy progression is governed entirely by the alignment between the state and the velocity:
\begin{equation}
\frac{d}{dt}E^{\mathrm{ODE}}_t = x_t^\top v(x_t,t)
\end{equation}
We recall that the reverse-time SDE drift relates to the ODE drift via the score function. The SDE trajectory is:
\begin{equation}
dx_t = \left[ v(x_t,t) - \frac{1}{2}g^2(t)s_\theta(x_t,t) \right] dt + g(t)\,d\bar{\mathrm{w}}
\end{equation}
To evaluate the energy differential of this stochastic process, we apply It\^o's lemma. Because the process is integrated backwards in time from $t=T$ to $t=0$, we define the backward time variable $\tau = T - t$, such that $d\tau = -dt > 0$. The forward-in-$\tau$ SDE is $dx_\tau = -\left[ v - \frac{1}{2}g^2s_\theta \right] d\tau + g\,d\mathrm{w}_\tau$. Applying It\^o's lemma for $f(x) = \frac{1}{2}\|x\|_2^2$ yields:
\begin{equation}
dE_\tau = x_\tau^\top dx_\tau + \frac{1}{2}\mathrm{Tr}\left[g^2(t)I \nabla_x^2 f(x_\tau)\right] d\tau = x_\tau^\top \left( -v + \frac{1}{2}g^2s_\theta \right) d\tau + \frac{D}{2}g^2 d\tau + x_\tau^\top g\,d\bar{\mathrm{w}}_\tau
\end{equation}
where $D$ is the data dimension. Reverting to the forward-time derivative $\frac{d}{dt} = -\frac{d}{d\tau}$ and taking the expectation to zero out the Wiener noise, the expected energy drift of the SDE is:
\begin{equation}
\frac{d}{dt}\mathbb{E}[E_t^{\mathrm{SDE}}] = \mathbb{E}[x_t^\top v(x_t,t)] - \frac{1}{2}g^2(t)\mathbb{E}[x_t^\top s_\theta(x_t,t)] - \frac{D}{2}g^2(t)
\end{equation}

\paragraph{The Ideal Score and Heat Cancellation.}
To understand this dynamic, assume an ideal oracle score $s_\theta = s^* := \nabla_{x_t}\log p_t(x_t)$. By integration by parts, the expected alignment of the data with its true score is strictly $\mathbb{E}[x_t^\top s^*] = -D$. Substituting this into the SDE drift:
\begin{equation}
\frac{d}{dt}\mathbb{E}[E_t^{\mathrm{SDE}}] = \mathbb{E}[x_t^\top v(x_t,t)] - \frac{1}{2}g^2(t)(-D) - \frac{D}{2}g^2(t) = \mathbb{E}[x_t^\top v(x_t,t)] = \frac{d}{dt}E_t^{\mathrm{ODE}}
\end{equation}
This mathematically formalizes the ideal generative balance: the stochastic heat explicitly injected by the It\^o noise term ($-\frac{D}{2}g^2$) is perfectly and exactly canceled by the restorative radial contraction of the true score ($\frac{D}{2}g^2$).

\paragraph{The Imperfect Score.}
In practice, the learned score contains approximation errors: $s_\theta(x_t,t) = s^*(x_t,t) + \epsilon(x_t,t)$. Substituting this imperfect score breaks the perfect heat cancellation:
\begin{equation}
\frac{d}{dt}\mathbb{E}[E_t^{\mathrm{SDE}}] = \frac{d}{dt}E_t^{\mathrm{ODE}} - \frac{1}{2}g^2(t)\mathbb{E}[x_t^\top \epsilon(x_t,t)]
\end{equation}
Integrating this drift difference from $t=T$ (where $E^{\mathrm{SDE}}_T = E^{\mathrm{ODE}}_T$) down to $t=0$, we isolate the exact energy gap between the generated distributions:
\begin{equation}
\mathbb{E}[E_0^{\mathrm{SDE}}] - E_0^{\mathrm{ODE}} = \int_T^0 -\frac{1}{2}g^2(t)\mathbb{E}[x_t^\top \epsilon(x_t,t)]\,dt = \int_0^T \frac{1}{2}g^2(t)\mathbb{E}[x_t^\top \epsilon(x_t,t)]\,dt
\end{equation}
Thus, the macroscopic divergence between SDE and ODE generation is controlled strictly by the state-error correlation term $\mathbb{E}[x_t^\top \epsilon(x_t,t)]$. If nonzero, the ``heat vs. contraction'' balance is broken, and pathwise energy drifts away from the ideal trajectory.

\subsubsection{State-Error Correlation and the Power Spectral Density Gap}
\label{app:subsubsec:state_error_cor_psd_gap}
By Parseval–Plancherel identity \citep{plancherel1910contribution}, this global correlation can be decomposed into independent frequency contributions:
\begin{equation}
x_t^\top \epsilon(x_t,t) = \sum_f \operatorname{Re}\left(\hat{x}_t(f)^*\,\hat{\epsilon}_t(f)\right)
\end{equation}
We formulate this explicitly for a single frequency $f$. Defining the correlation functional as $\Gamma_f(t) := \mathbb{E}\left[\operatorname{Re}\left(\hat{x}_t(f)^*\,\hat{\epsilon}_t(f)\right)\right]$, the band-wise energy gap at $t=0$ obeys:
\begin{equation}
\label{eq:sign_state_err_cor}
\Delta_f := \mathbb{E}\big[||\hat{x}_0^{\mathrm{SDE}}(f)||^2\big] - \mathbb{E}\big[||\hat{x}_0^{\mathrm{ODE}}(f)||^2\big] = \int_0^T g^2(t)\Gamma_f(t)\,dt
\end{equation}
Because $g^2(t) \ge 0$, the sign of $\Delta_f$ is determined exclusively by the accumulated sign of $\Gamma_f(t)$. To determine whether the network error $\epsilon$ induces positive or negative state-error correlation, we must account for neural network learning dynamics. Networks trained via Mean Squared Error (MSE) exhibit regression to the mean; when confronted with uncertainty, they output a conservative, smoothed estimate \cite{pmlr-v48-larsen16}. Consequently, $s_\theta$ systematically underestimates the magnitude of the true score $s^*$. We model this approximation error as being anti-aligned with the true score: $\epsilon_t \approx -\alpha_f s^*_t$ where $0 < \alpha_f < 1$.

To determine the direction of the true score $s^*$, we invoke Tweedie's formula \citep{efron2011tweedie, chung2022diffusion}, which states that the score is proportional to the difference between the expected clean data and the current noisy state: $\hat{s}^*(f, t) \propto (1-t)\mathbb{E}[\hat{x}_0(f) \mid x_t] - \hat{x}_t(f)$. Let $N_f$ represent the energy of the initial noise prior, and $R_f$ represent the target energy of the real data distribution. To ground this physically, natural images exhibit a well-documented $1/f^\omega$ power-law spectrum \citep{VANDERSCHAAF19962759, hyvärinen2009natural}, a structural property that fundamentally persists even within the compressed latent spaces of modern autoencoders \citep{ning2026spectrum}. Thus, we can approximate the real-data spectral target as a normalized $R_f \approx C/f^{\omega_\text{pow}}$ (with $\omega_\text{pow} \approx 2$), while the standard Gaussian prior maintains a flat white-noise spectrum $N_f = 1$. Comparing these frequency-dependent magnitudes defines three distinct generative regimes:
\subparagraph{1. The Attenuation Regime ($R_f < N_f$):}
At frequencies where the target data energy is lower than the initial noise (typically high frequencies), the required evolution is attenuation. Because the expected clean signal magnitude is smaller than the current noisy state, the vector difference points inward. Thus, the true score acts to destroy noise: $\hat{s}^* \propto -c_1\hat{x}_t$. Because the network underestimates this inward pull, the resulting error points outward ($\hat{\epsilon}_t \propto +\alpha_f c_1\hat{x}_t$). The error is positively aligned with the state, meaning their inner product is positive:
\begin{equation}
\Gamma_f(t) > 0 \implies \Delta_f > 0
\end{equation}
\textbf{Conclusion:} The SDE over-allocates energy relative to the ODE. The continuous noise injection is not fully dissipated because the learned score is too weak to adequately attenuate the state.

\subparagraph{2. The Amplification Regime ($R_f > N_f$):}
At frequencies where the target structural magnitude is larger than the initial noise (typically low frequencies), the required evolution is amplification. The expected clean data vector has a significantly larger magnitude ($\|\mathbb{E}[\hat{x}_0(f)]\| > \|\hat{x}_t(f)\|$). Thus, the vector difference points outward: $\hat{s}^* \propto c_2\hat{x}_t$. The underestimation error therefore points inward ($\hat{\epsilon}_t \propto -\alpha_f c_2\hat{x}_t$). The error is anti-aligned with the state, yielding a negative inner product:
\begin{equation}
\Gamma_f(t) < 0 \implies \Delta_f < 0
\end{equation}
\textbf{Conclusion:} The SDE under-allocates energy relative to the ODE. The SDE's constant stochastic disruption, coupled with a weakened score that cannot fully drive the necessary amplification, causes the trajectory to fall short of the ideal energy level.

\subparagraph{3. The Crossover Point ($R_f = N_f$):}
The regime transition occurs exactly at the frequency where the inherent energy of the initial noise matches the target energy of the real data. Here, the expected magnitude of the clean data equals the magnitude of the current noisy state. The score provides a purely tangential (phase-rotational) pull, exerting zero radial pull. Consequently, the inner product of the state and the score (and therefore the error) is zero:
\begin{equation}
\Gamma_f(t) = 0 \implies \Delta_f = 0
\end{equation}
\textbf{Conclusion:} At the exact crossover frequency, the radial approximation error vanishes, and the ODE and SDE energy trajectories perfectly match.

\subsection{Spectral Impact and Energy Dynamics of Colored Noise SDEs}
\label{app:subsec:cns_spectral_impact}
As proven in App.~\ref{app:subsec:invariance}, the total energy injected by the stochastic process is finite and strictly bounded by the diffusion schedule $g(t)$. We now seek to mathematically formalize how a Colored Noise SDEs reshapes the continuous evolution of this energy on a strictly band-wise basis.

\subsubsection{Energy Dynamics under an Ideal Score Function}

\paragraph{Uniform Energy Allocation in Standard SDEs.}
In a standard white-noise SDE, the injected stochastic power is distributed uniformly across all frequency bands at every timestep. Equivalently, each integration step injects a flat Power Spectral Density (PSD), ensuring all spatial frequencies receive an equal share of the finite injected energy budget.

\paragraph{The Colored Noise Variance Operator.}
We introduce a colored-noise process $d\widetilde{\mathrm{w}}_t$ by scaling the standard noise differently across frequency bands. In the Fourier domain, let:
\begin{equation}
B(t) = \operatorname{diag}(\beta_1(t), \beta_2(t), \dots, \beta_D(t))
\end{equation}
where each $\beta_f(t) > 0$ is a real, time-dependent scaling weight for frequency index $f$, and $D$ is the data dimensionality. The modified noise increment in the frequency domain is defined as:
\begin{equation}
d\widehat{\widetilde{\mathrm{w}}}_t = B(t)\,d\widehat{\bar{\mathrm{w}}}_t, \qquad d\widehat{\widetilde{\mathrm{w}}}_t(f) = \beta_f(t)\,d\widehat{\bar{\mathrm{w}}}_t(f)
\end{equation}
Mapping back to the spatial domain via the inverse Fourier transform $\mathcal{F}^{-1}$ yields the linear operator form:
\begin{equation}
d\widetilde{\mathrm{w}}_t = \mathcal{F}^{-1} B(t) \mathcal{F}\, d\bar{\mathrm{w}}_t
\end{equation}

\paragraph{The Global Variance-Conservation Constraint.}
To preserve the global stability of the generative process (shown necessity in App.~\ref{app:subsec:variance_conservation}), we require the total injected energy of CNS to precisely match the standard white-noise SDE. For our modified process, the expected variance is:
\begin{equation}
\mathbb{E}\left[||g(t)\,d\widetilde{\mathrm{w}}_t||^2\right] = g^2(t)\sum_{f=1}^{D} \beta_f^2(t)\,\mathbb{E}\left[||d\widehat{\bar{\mathrm{w}}}_t(f)||^2\right] = g^2(t) \sum_{f=1}^{D} \beta_f^2(t)\,dt
\end{equation}
Because the standard SDE injects a total variance of $g^2(t) \cdot D \cdot dt$, matching this global energy budget imposes a strict variance-conservation constraint (i.e., the weights must have a Root Mean Square of 1):
\begin{equation}
\frac{1}{D}\sum_{f=1}^{D}\beta_f^2(t) = 1
\end{equation}

\paragraph{Band-Wise Energy Drift of Colored Noise SDE.}
Substituting this scaled noise process into the reverse-time SDE dynamics, the Fourier-domain state evolution for a specific frequency band $f$ becomes:
\begin{equation}
d\widehat{x}_t(f) = \widehat{\mu}(x_t,t)_f\,dt + g(t)\beta_f(t)\,d\widehat{\bar{\mathrm{w}}}_t(f)
\end{equation}
Applying It\^o's lemma to the band energy $E_t^{(f)} := \frac{1}{2}||\widehat{x}_t(f)||^2$, as derived in App.~\ref{app:subsec:spectral_gap_origins}, the expected band-wise energy drift under the CNS takes the form:
\begin{equation}
\label{eq:CNSDE_energy_evolution}
\frac{d}{dt}\mathbb{E}\left[E_t^{(f),\mathrm{CNS}}\right] = \mathbb{E}\left[\operatorname{Re}\left(\widehat{x}_t(f)^*\widehat{\mu}(x_t,t)_f\right)\right] - \frac{1}{2}g^2(t)\beta_f^2(t)
\end{equation}  For comparison, under the standard white-noise SDE ($\beta_f(t) \equiv 1$), the corresponding band-wise drift is:
\begin{equation}
\frac{d}{dt}\mathbb{E}\left[E_t^{(f),\mathrm{SDE}}\right] = \mathbb{E}\left[\operatorname{Re}\left(\widehat{x}_t(f)^*\widehat{\mu}(x_t,t)_f\right)\right] - \frac{1}{2}g^2(t)
\end{equation}
Hence, the colored-noise modification dynamically alters only the injected It\^o heat term from $\frac{1}{2}g^2(t)$ to $\frac{1}{2}g^2(t)\beta_f^2(t)$, enabling targeted frequency-dependent amplification ($\beta_f^2 > 1$) or attenuation ($\beta_f^2 < 1$) while safely preserving the global energy budget.

\textbf{Interpretation and Required Assumptions.}
For theoretical tractability (while noting this is an approximation due to the network's spectral bias), assume the ideal oracle score applies a per-band restoring drift that perfectly cancels the standard It\^o energy injection. See App.~\ref{app:state_in_distribution} for further details.  Under these assumptions, the perfect score cancels each band at the standard unit scale. Consequently, scaling the injected noise in band $f$ by $\beta_f(t)$ creates an intentional mismatch: $\beta_f^2 > 1$ yields a net energy addition in that band, while $\beta_f^2 < 1$ yields a net energy dissipation. Assuming the model reliably maps this altered stochastic energy into coherent spatial structures (detailed in Sec.~\ref{sec:energy_transfer}), higher injected variance reliably yields higher generated structural energy in that frequency band.  \subsubsection{Zero-Sum Spectral Constraints under an Ideal Score}We isolate the explicit effect of noise shaping by subtracting the standard SDE energy drift from the CNS drift, yielding the instantaneous excess energy injected into band $f$:
\begin{equation}
\frac{d}{dt}\left(\mathbb{E}\left[E_t^{(f),\mathrm{SDE}}\right] - \mathbb{E}\left[E_t^{(f),\mathrm{CNS}}\right]\right) = \frac{1}{2}g^2(t)\left(\beta_f^2(t) - 1\right)
\end{equation}
Integrating this over the sampling trajectory $t \in [T, 0]$ formalizes the terminal excess-energy gap for frequency $f$:
\begin{equation}
\mathrm{Excess}_f := \int_{0}^{T} \frac{1}{2}g^2(t)\sum_f\left(\beta_f^2(t) - 1\right)dt
\end{equation}
\textbf{Conclusion (Zero-Sum Redistribution).}
If we scale a specific frequency band such that $\beta_f > 1$, the generated energy in that band increases. However, due to the global conservation constraint ($\frac{1}{D}\sum \beta_f^2 = 1$), deviations from the unit baseline must sum strictly to zero:
\begin{equation}
\sum_{f=1}^{D}\left(\beta_f^2(t) - 1\right) = 0
\end{equation}
Therefore, summing the excess energies over all bands confirms global conservation:
\begin{equation}
\sum_{f=1}^{D}\mathrm{Excess}_f = \int_{0}^{T} \frac{1}{2}g^2(t)\left(\sum_{f=1}^{D}(\beta_f^2(t) - 1)\right) \,dt = 0
\end{equation}
Under an ideal score, any spectral adjustment is fundamentally a zero-sum game: specific bands may achieve higher energy under CNS, but exclusively at the expense of others.

\subsubsection{Breaking the Zero-Sum Constraint via Score Approximation Error}
The zero-sum paradigm relies on a critical assumption: that the conversion rate of raw injected noise into permanent generated structural energy is constant across all frequency bands. However, as derived in App.~\ref{app:subsec:spectral_gap_origins}, the actual amount of injected raw noise that successfully converts into permanent structural divergence is strictly dictated by the state--error correlation term $\Gamma_f(t)$. We now demonstrate that this correlation is highly non-stationary and decays rapidly as the structural content of the image resolves. To quantify structural formation, we use the normalized bounded metric representing how ``built'' a specific frequency band $f$ is at any time $t$ derived in Sec.~\ref{sec:spectral_bias} and visually shown in Fig.~\ref{fig:gamma_matrix}.

\paragraph{Proposed Mechanisms for State-Error Correlation Decay}
\label{app:correlation_decay_mechanisms}
To theoretically explain the decay in state-error correlation ($\lim_{\gamma_f(t) \to 1} \Gamma_f(t) = 0$), we postulate two complementary mechanisms: (1) the leading-order radial component of the true score vanishes as a band resolves, and (2) a possible transition of the network's error on unresolved high-frequency details toward phase-random scatter across consecutive late-phase steps. 

We note that the framework presented here is intended as a plausibility argument rather than a rigorous derivation. In particular, we do not require the absolute magnitude of the score error $\|\epsilon\|$ to vanish; we merely suggest that the changing geometric and temporal properties of the terminal generative state could plausibly drive the accumulated correlation toward zero even while $\|\epsilon\| \gg 0$.

\textbf{1. Tangential Dominance of the True Score.}
By Tweedie's formula \citep{efron2011tweedie}, the true score in the frequency domain is proportional to the displacement from the current state toward its conditional clean estimate:
\begin{equation}
\hat{s}^*(f, t) \;\propto\; \mathbb{E}[\hat{x}_0(f) \mid x_t] - \hat{x}_t(f)
\end{equation}
We suppress the schedule-dependent prefactor here because only the \emph{direction} of $\hat{s}^*$ relative to $\hat{x}_t$ enters the radial inner product we are about to compute. When a frequency band approaches a fully ``built'' state ($\gamma_f(t)\to 1$), the network has correctly resolved the band's macroscopic content, so the conditional clean estimate is closely aligned with the current state in both magnitude and approximate direction. We characterize the residual mismatch as a small phase rotation:
\begin{equation}
\mathbb{E}[\hat{x}_0(f) \mid x_t]
\;\approx\; e^{i\,\delta\phi_f(t)}\,\hat{x}_t(f),
\qquad \delta\phi_f(t)\to 0 \;\text{ as }\; \gamma_f(t)\to 1.
\end{equation}
This is strictly stronger than merely $\|\mathbb{E}[\hat{x}_0(f)\mid x_t]\| \approx \|\hat{x}_t(f)\|$: equal magnitudes alone leave room for an arbitrary chord-like displacement with a substantial radial component, whereas a small-angle phase rotation is tangential to leading order. Expanding:
\begin{equation}
\hat{s}^*(f,t) \;\propto\; \left(e^{i\delta\phi_f}-1\right)\hat{x}_t(f)
\;=\; i\,\delta\phi_f\,\hat{x}_t(f) \;+\; \mathcal{O}(\delta\phi_f^2).
\end{equation}
The leading-order displacement $i\,\delta\phi_f\,\hat{x}_t$ is geometrically orthogonal to $\hat{x}_t$ in the complex plane (multiplication by $i$ rotates by $90^\circ$), so the radial inner product of state and score is purely a quadratic remainder:
\begin{equation}
\operatorname{Re}\!\left(\hat{x}_t(f)^*\,\hat{s}^*(f,t)\right)
\;=\; \mathcal{O}(\delta\phi_f^2).
\end{equation}
Carrying the early-phase underestimation model $\hat{\epsilon}_t \approx -\alpha_f \hat{s}^*$ from App.~\ref{app:subsec:spectral_gap_origins} forward, the same quadratic suppression propagates to the state-error correlation:
\begin{equation}
\operatorname{Re}\!\left(\hat{x}_t(f)^*\,\hat{\epsilon}_t(f)\right)
\;\xrightarrow{\;\gamma_f(t)\to 1\;}\; 0.
\end{equation}

\textbf{2. Transition to Phase-Random Error on Unresolved Details.}
During the early phases of band formation ($\gamma_f(t) \ll 1$), MSE training induces a temporally coherent radial bias—specifically, a systematic underestimation of the score's amplitude along the state direction. However, this coherence breaks once the band's macroscopic magnitude is established ($\gamma_f(t) \to 1$). At this terminal stage, the network's residual task shifts to resolving fine sub-structural details, such as sharp edges and exact phase alignments. Because these high-frequency features are difficult to infer from a condensed latent representation, the network's errors abandon any consistent directional bias, transitioning instead into erratic, phase-random fluctuations across consecutive late-phase integration steps.

To see how such a transition could matter, it is useful to write the band-wise inner product in polar form:
\begin{equation}
\operatorname{Re}\!\left(\hat{x}_t(f)^*\,\hat{\epsilon}_t(f)\right)
= |\hat{x}_t(f)|\,|\hat{\epsilon}_t(f)|\,\cos\!\left(\theta_\epsilon(t)
  - \theta_x(t)\right).
\end{equation}
The contribution of the late steps to the accumulated energy gap is governed by the time average of $\cos(\theta_\epsilon - \theta_x)$ along their trajectory. In the early phase this relative phase is presumably concentrated near $0$ or $\pi$ for many consecutive steps (the radial directions associated with the underestimation model), so per-step contributions would add coherently with a fixed sign. If, in the terminal regime, the relative phase instead varied erratically across the final taken steps and behaved roughly as if uniform on $[0,2\pi)$, the signed per-step contributions would tend to cancel under temporal accumulation, and the late-phase contribution to $\Gamma_f$ could plausibly collapse toward zero --- even while each individual step still satisfied $\|\hat{\epsilon}_t(f)\|\gg 0$. We offer this as a speculative mechanism that, together with the geometric argument above, may help account for the proposed decay of $\Gamma_f$.

\paragraph{State-Dependent Efficiency and the Breakdown of Zero-Sum Redistribution.}
Together, these proposed mechanisms provide a theoretical framework explaining why the state-error correlation $\Gamma_f(t) = \mathbb{E}\left[\operatorname{Re}\left(\hat{x}_t(f)^*\hat{\epsilon}_t(f)\right)\right]$ collapses once a frequency band is structurally resolved. Under this framework, when a band is fully built ($\Gamma_f(t) \approx 0$), the SDE manages to properly balance stochastic noise injection and score-driven contraction.
Here, the same $\Gamma_f$ that governs the band-wise SDE-vs-ODE gap (Eq.~\ref{eq:sign_state_err_cor}) acts as a \emph{conversion gain}. It quantifies the rate at which raw injected stochastic variance is translated into coherent radial growth, rather than dissipating as incoherent fluctuation. As $\Gamma_f \to 0$, this conversion gain vanishes. 
Consequently, any surplus variance ($\beta_f^2 > 1$) enters the instantaneous variance budget but fails to materialize as coherent image content; instead, subsequent denoising steps treat it as ambient noise and remove it. As a result, the global conservation of injected variance ($\sum (\beta_f^2 - 1) = 0$) does not enforce a zero-sum conservation of \emph{generated} energy, effectively dismantling the zero-sum redistribution constraint.

Ultimately, the true efficacy of colored noise---its \emph{energy absorbance efficiency}---is dictated by the temporal intersection of the variance scaling $\beta_f(t)$ and the decaying structural correlation $\Gamma_f(t)$. Permanently altering the generated spectrum requires injecting variance while a band remains in structural deficit ($\gamma_f(t) \ll 1$). This highlights a fundamental property of stochastic generation: macroscopic spectral divergence is not determined by the total volume of injected energy, but by its timing.

\paragraph{Derivation of the Proposed CNS Allocation Schedule.}
\label{app:colored_noise_generated_spectrum}
Having established that energy conversion efficiency decays as $\gamma_f(t) \to 1$, we formulate the optimal time-dependent injection strategy. To maximize the utility of the finite variance budget, the system must avoid injecting excess energy into ``built'' bands where it will be wastefully dissipated. Instead, it must dynamically route variance into ``unbuilt'' lagging frequencies where the correlation $\Gamma_f(t)$ remains high, maximizing the structural retention of the injected heat.  While a naive maximization strategy might suggest a ``bang-bang'' controller \citep{liberzon2011calculus, kamien2013dynamic}---routing the entire variance budget exclusively to the single least-built frequency at any given moment---we hypothesize that energy conversion efficiency is fundamentally bounded by the local magnitude of the injected noise. Overwhelming a single frequency band with excessive stochastic variance surpasses the local restorative capacity of the score network, degrading the model's ability to cohesively integrate or dissipate that noise. Consequently, the optimal schedule must avoid greedy allocation. The variance scaling factor $\beta_f^2(t)$ should instead be smoothly and directly proportional to the structural deficit of the frequency band, $(1 - \gamma_f(t))$. To satisfy the global energy conservation constraint ($\frac{1}{D}\sum \beta_f^2(t) = 1$) at every timestep, we normalize this deficit across the frequency spectrum by its Root Mean Square (RMS):
\begin{equation}
\beta_f(t) = \frac{\sqrt{1-\gamma_f(t)}}{\sqrt{\frac{1}{D}\sum_{f'=1}^{D} \left(1-\gamma_{f'}(t)\right)}}
\end{equation}  This schedule exhibits several highly desirable theoretical properties:
\begin{itemize}
\item \textbf{Initialization as White Noise:} At $t = T$, all bands are entirely unbuilt ($\gamma_f(T) = 0$). The allocation evaluates to $\beta_f(T) = 1$ for all frequencies. The generative process naturally begins as a standard uniform white-noise SDE.
\item \textbf{Dynamic Routing:} As generation progresses, bands accumulate structure at staggered rates. The schedule autonomously drains the variance budget from bands approaching $\gamma_f \approx 1$ and actively routes it into lagging frequencies.
\item \textbf{Avoidance of Local Saturation:} By scaling proportionally rather than utilizing a binary ``bang-bang'' injection, the schedule prevents any single frequency band from being overwhelmed by stochastic variance, ensuring the noise remains within the operable restorative capacity of the score network.
\item \textbf{Optimal Retention:} By shifting the mathematical weight exclusively to regions where the state--error correlation is strongest, the schedule mathematically circumvents the limitations of the zero-sum energy game, ensuring that every unit of injected variance is optimally converted into permanent macroscopic structure.
\end{itemize}

\section{Methodological and Experimental Details}
\label{app:implementation_details}

This appendix provides the comprehensive methodological and experimental details required to reproduce the Colored Noise Sampling (CNS) framework. While the main text establishes the theoretical premise of spectral energy control, this section translates these concepts into a robust and readily deployable inference algorithm. First, in App.~\ref{app:subsec:radial_frequency_usage}, we formally define the multi-dimensional Fourier projection operators standardly used to isolate and evaluate isotropic radial frequency bands. Next, in App.~\ref{app:empirical_relaxations}, we detail our hardware configuration alongside practical schedule relaxations---such as progression scaling and dynamic spectral tilting---that seamlessly integrate CNS into standard sampling pipelines, further enhance algorithmic robustness, and consistently improve synthesis results.

\subsection{Isolation of Radial Spatial Frequency Bands}
\label{app:subsec:radial_frequency_usage}

Let $x \in \mathbb{R}^{C \times H \times W}$ denote a continuous, real-valued multi-channel spatial tensor, such as a generated noise sample or an image. To accurately evaluate the frequency-coupled dynamics of the diffusion process, we must systematically isolate the signal components corresponding to specific spatial scales. This is achieved using the multi-dimensional discrete Fourier transform.

\paragraph{The 2D Discrete Fourier Transform and Shifted Grid.}
We begin by applying the 2D Discrete Fourier Transform (DFT), denoted as $\mathcal{F}$, independently across each channel $c$:
\begin{equation}
    X_c(u, v) = \mathcal{F}[x_c](u, v) = \sum_{h=0}^{H-1} \sum_{w=0}^{W-1} x_c(h, w) e^{-i 2\pi \left(\frac{u h}{H} + \frac{v w}{W}\right)}
\end{equation}
To analyze frequencies based on their spatial scale rather than their directional orientation, we shift the 2D frequency indices to center the DC (zero-frequency) component at the origin. Let the shifted frequency coordinates be denoted as $(f_y, f_x)$, defined over the discrete domain:
\begin{equation}
    f_y \in \left[-\lfloor H/2 \rfloor, \dots, \lceil H/2 \rceil - 1\right], \quad f_x \in \left[-\lfloor W/2 \rfloor, \dots, \lceil W/2 \rceil - 1\right]
\end{equation}

\paragraph{Isotropic Radial Frequencies.}
Because natural images and standard diffusion noise priors are generally isotropic (rotationally invariant in expectation), we collapse the 2D frequency grid into a 1D measure of spatial scale. We define the radial frequency $\rho(f_y, f_x)$ as the Euclidean distance from the DC component:
\begin{equation}
    \rho(f_y, f_x) = \sqrt{f_y^2 + f_x^2}
\end{equation}
The maximum theoretical radial frequency, corresponding to the corners of the 2D Nyquist limit, is bounded by $\rho_{\max} = \sqrt{(H/2)^2 + (W/2)^2}$.

\paragraph{Discrete Band Partitioning.}
To perform statistical analysis across the spectrum, we partition the continuous range $[0, \rho_{\max}]$ into $N_b$ discrete radial frequency bands. Each discrete coordinate $(f_y, f_x)$ is mapped to an integer band index $b \in \{0, \dots, N_b - 1\}$ via nearest-integer scaling:
\begin{equation}
    b(f_y, f_x) = \left\lfloor \frac{\rho(f_y, f_x)}{\rho_{\max}} (N_b - 1) \right\rceil
\end{equation}
This mapping defines a set of mutually exclusive 2D frequency masks, $B_b$, where each set contains all coordinate pairs belonging to band $b$:
\begin{equation}
    B_b = \left\{ (f_y, f_x) \in \mathbb{Z}^2 : b(f_y, f_x) = b \right\}
\end{equation}

\paragraph{The Projection Operator and Hermitian Symmetry.}
Using these discrete sets, we define the band-pass projection operator $P_b[\cdot]$, which isolates the spatial signal residing exclusively in band $b$. This is achieved by taking the inverse 2D DFT of the masked spectrum:
\begin{equation}
    P_b[x] = \mathcal{F}^{-1} \left[ \mathbf{1}_{(f_y, f_x) \in B_b} \odot \mathcal{F}[x] \right]
\end{equation}
where $\mathbf{1}$ is the indicator function and $\odot$ denotes element-wise multiplication. 

A critical geometric property of the radial distance function $\rho$ is its symmetry across the origin; if $(f_y, f_x) \in B_b$, then $(-f_y, -f_x) \in B_b$. Because the input spatial tensor $x$ is real-valued, its Fourier transform inherently exhibits Hermitian conjugate symmetry ($X(-f_y, -f_x) = X^*(f_y, f_x)$). The isotropic symmetry of the mask $B_b$ perfectly preserves this Hermitian property during the masking operation. Consequently, the inverse DFT guarantees that the projected tensor remains strictly real-valued:
\begin{equation}
    P_b[x] \in \mathbb{R}^{C \times H \times W}
\end{equation}
This allows us to treat $P_b[x]$ not as a complex Fourier series, but as a standard real vector in a lower-dimensional subspace $\mathcal{V}_b \subset \mathbb{R}^{CHW}$. Because the projection operator returns real values, we can seamlessly compute standard spatial metrics---such as the $L_2$ norm and cosine similarity---directly on the projected tensors.
\subsection{Hardware Configuration and Empirical Relaxations}
\label{app:empirical_relaxations}

\paragraph{Hardware and Environment.} 
All generative inference and evaluation experiments were conducted using PyTorch with Distributed Data Parallel (DDP) across a compute node equipped with four NVIDIA L40S (48GB VRAM) GPUs.

\paragraph{Empirical Schedule Relaxations.} 
While App.~\ref{app:colored_noise_generated_spectrum} derives the theoretically optimal $\beta(f,t)$ variance allocation schedule, we empirically found that introducing minor algorithmic relaxations helps maintain optimal generative stability. In practice, we implement three parameterized adjustments to our inference framework:

\begin{itemize}
    \item \textbf{Progression Scaling:} To prevent premature variance routing, we soften the strictly computed structural progression matrix by introducing a constant scaling factor $c > 1$, such that the effective progression becomes $\tilde{\gamma}(f,t) = \gamma(f,t) / c$.
    \item \textbf{Dynamic Spectral Tilting:} For specific experiments, we apply a temporally evolving exponential tilt to the injected colored noise spectrum. This smooths the strict band-wise allocation, providing a more stable transition across adjacent frequency bands as generation progresses.
    \item \textbf{Energy Equilibrium Tuning:} Although App.~\ref{app:subsec:variance_conservation} derives a strict $\beta^2 = 1$ constraint, empirical score networks systematically underestimate the restorative gradient (App.~\ref{app:correlation_decay_mechanisms}), skewing the reverse-time SDE's ``heat vs. contraction'' balance. We compensate with a micro-scaling of the total injected energy budget; the values used ($0.98$--$0.999$, Tab.~\ref{tab:supp_hyperparams}) lie well inside the unit-energy stability basin characterized in Tab.~\ref{tab:ablation_studies_appendix}.
\end{itemize}

Exact configuration values and hyperparameter selections across all evaluated architectures (SiT, JiT, FLUX) are comprehensively reported in Tab.~\ref{tab:supp_hyperparams}.

\begin{table}[h]
\centering
\caption{Sampling hyperparameters for the baseline architectures evaluated in our experimental setup. Standard experiments.}
\footnotesize
\setlength{\tabcolsep}{3.5pt} 
\begin{tabular}{l cc ccc cc}
\toprule
& \multicolumn{2}{c}{\textbf{SiT-XL/2}} & \multicolumn{3}{c}{\textbf{JiT-B/16}} & \multicolumn{2}{c}{\textbf{JiT-H/16}} \\
\cmidrule(lr){2-3} \cmidrule(lr){4-6} \cmidrule(lr){7-8}
\textbf{Parameter} & \textbf{w/o CFG} & \textbf{w/ CFG} & \textbf{w/o CFG} & \textbf{w/ CFG} & \textbf{w/ CFG} & \textbf{w/o CFG} & \textbf{w/ CFG} \\
\midrule
\rowcolor[gray]{0.95} \multicolumn{8}{l}{\textit{Architecture \& Data}} \\
Space & \multicolumn{2}{c}{Latent (VAE)} & \multicolumn{3}{c}{Pixel} & \multicolumn{2}{c}{Pixel} \\
Dataset & \multicolumn{2}{c}{ImageNet-256} & \multicolumn{3}{c}{ImageNet-256} & \multicolumn{2}{c}{ImageNet-256} \\
Prediction & \multicolumn{2}{c}{$v$-pred} & \multicolumn{3}{c}{$x$-pred} & \multicolumn{2}{c}{$x$-pred} \\
Frequency Bands & \multicolumn{2}{c}{32} & \multicolumn{3}{c}{32} & \multicolumn{2}{c}{32}\\
Sampling Steps & \multicolumn{2}{c}{250} & \multicolumn{3}{c}{50} & \multicolumn{2}{c}{50}\\

\midrule
\rowcolor[gray]{0.95} \multicolumn{8}{l}{\textit{CNS Sampling Settings}} \\
Guidance Scale & -- & 1.5 & -- & 2.6 & 3.0 & -- & 2.2 \\
Solver & Euler/Heun & Euler & Euler & Euler & Euler & Euler & Euler \\
$\gamma(t,f)$ Power & 0.75 & 0.5 & 0.5 & 0.5 & 0.5 & 0.5 & 0.5 \\
$\gamma(f,t)$ Divider & 1.73 & 25.0 & 5.0 & 10.0 & 14.0 & 7.5 & 25.0 \\
Alpha Tilting Interpolation & $(0.15,-0.5)$ & $(-0.1, 0.03)$ & - & - & - & - & - \\
Alpha Tilting Interpolation Type & Exponential ($0.75$) & Linear & - & - & - & - & - \\
Noise Energy Scale & 0.98 & 0.998 & 0.995 & - & - & 0.98 & 0.999 \\
\bottomrule
\end{tabular}
\label{tab:supp_hyperparams}
\end{table}

\section{Additional Results and Extended Evaluations}
\label{sec:appendix_additional_results}

This section provides extended empirical evaluations that supplement the findings presented in the main text. We first present additional quantitative benchmarks across varying architectures and sampling settings (App.~\ref{app:extended_benchmarks}). Next, we detail the numerical integration schemes utilized for our stochastic differential equations high-order solvers (App.~\ref{app:solver_details}) and analyze the robustness of CNS across varying numbers of discretization steps (App.~\ref{app:discretization_details}). To rigorously validate our theoretical constraints and design choices, we then provide a comprehensive ablation study evaluating the effects of global energy scaling, spectral perturbation, and temporal schedule misalignment (App.~\ref{app:subsec:ablation_study_full}). Finally, we elaborate on experiments demonstrating the orthogonality and generalization of CNS to models trained with alternative noise distributions (App.~\ref{app:bndm_details}). Finally, we conclude with extended qualitative visual comparisons (without guidance) that highlight the superiority of CNS over standard baselines (App.~\ref{sec:appendix_Visual_compare}).

\subsection{Extended Generative Benchmarks}
\label{app:extended_benchmarks}
To demonstrate consistency across model architectures and evaluation frameworks, we provide supplementary performance tables. Tab.~\ref{tab:jitH16-100-steps-evaluation-table} reports the unguided JiT-H/16 results utilizing 100 sampling steps. Under these conditions, CNS strictly dominates both the ODE and standard SDE baselines across all tracked metrics. Furthermore, Tab.~\ref{tab:geneval_flux_results} (referenced in the main text) details the extensive GenEval \citep{ghosh2023geneval} benchmarking for the FLUX model, demonstrating CNS's capacity to enhance complex text-to-image synthesis.
\begin{table}[h]
  \caption{\textbf{Evaluation of Unguided Image Generation Performing 100 Sampling Steps}. ImageNet-256 JiT-H/16 model evaluation metrics without Classifier-Free Guidance across different sampling methods.}
  \label{tab:jitH16-100-steps-evaluation-table}
  \centering
  \begin{tabular}{lcccccc}
    \toprule
    Sampler & FID $\downarrow$ & sFID $\downarrow$ & IS $\uparrow$ & Prec. $\uparrow$ & Rec. $\uparrow$ \\
    \midrule
    ODE & 11.46 & 10.49 & 44.71 & 0.63 & 0.64 \\
    SDE & 8.95 & 7.98 & 46.38 & 0.65 & \textbf{0.65} \\
    CNS (Ours) & \textbf{8.57} & \textbf{7.16} & \textbf{46.72} & \textbf{0.66} & \textbf{0.65} \\
    \bottomrule
  \end{tabular}
\end{table}

\begin{table}[h]
  \centering
  \caption{\textbf{GenEval quantitative compositional evaluation on FLUX.1-dev.} All results are reported as accuracy scores $\in [0, 1]$ (higher is better).}
  \label{tab:geneval_flux_results}
  \footnotesize
  \begin{tabularx}{\textwidth}{X c c c c c c c}
    \toprule
    \multirow{2}{*}{\textbf{Model \& Sampler}} & \multirow{2}{*}{\textbf{Overall}} & \multicolumn{6}{c}{\textbf{GenEval Task Breakdown}} \\
    \cmidrule(lr){3-8}
    & & \textbf{Single Obj.} & \textbf{Two Obj.} & \textbf{Counting} & \textbf{Colors} & \textbf{Color Attr.} & \textbf{Position} \\
    \midrule
    \multicolumn{8}{c}{\textit{FLUX.1-dev}} \\
    \midrule
    ODE & 0.643 & \textbf{0.988} & 0.784 & 0.699 & \textbf{0.826} & 0.413 & 0.152 \\
    SDE & 0.635 & 0.984 & 0.789 & 0.685 & 0.768 & \textbf{0.433} & 0.154 \\
    CNS (Ours) & \textbf{0.647} & \textbf{0.988} & \textbf{0.794} & \textbf{0.714} & 0.803 & 0.420 & \textbf{0.164} \\
    \bottomrule
  \end{tabularx}
\end{table}

\subsection{Numerical Integration Schemes for SDEs}
\label{app:solver_details}

When discretizing the reverse-time Stochastic Differential Equation (SDE) of a generative model, $d\mathbf{x} = \mathbf{f}(\mathbf{x}, t)dt + \mathbf{g}(\mathbf{x}, t)d\mathbf{w}$ \citep{song2020score}, the choice of numerical solver dictates the approximation error. Unlike Ordinary Differential Equations (ODEs) which rely on standard Taylor series approximations, SDEs require It\^{o}-Taylor expansions \citep{kloeden2011numerical}. Consequently, SDE solvers are characterized by two distinct types of convergence orders:

\begin{itemize}
    \item \textbf{Strong Order (Pathwise Accuracy):} A solver has strong order $p$ if the expected error of a single trajectory scales as $\mathcal{O}(h^p)$ as the step size $h \to 0$, defined as $\mathbb{E}[\|\mathbf{x}_N - \mathbf{x}(T)\|] \le C h^p$. This measures how accurately the solver tracks a specific noise realization.
    \item \textbf{Weak Order (Distributional Accuracy):} A solver has weak order $q$ if the error in the expectation of smooth test functions $\phi$ scales as $\mathcal{O}(h^q)$, defined as $|\mathbb{E}[\phi(\mathbf{x}_N)] - \mathbb{E}[\phi(\mathbf{x}(T))]| \le C h^q$. 
\end{itemize}

In the context of generative modeling, our primary objective is to sample from the correct target distribution rather than accurately track a specific microscopic noise path. Because standard evaluation metrics like Fréchet Inception Distance (FID) \citep{heusel2017gans} measure distributional distances, the \textit{weak} order of the solver is the dominant quantity of interest. In our experiments (Sec.~\ref{subsec:experiments_C2I}), we evaluate solvers of varying weak orders:

\begin{enumerate}
    \item \textbf{Euler-Maruyama} \citep{maruyama1955continuous}: The foundational 1st-order weak (and 1/2-order strong) SDE solver, requiring 1 function evaluation per step.
    \item \textbf{Stochastic Heun}: A 2nd-order weak predictor-corrector method requiring 2 function evaluations per step \citep{heun1900neue, karras2022elucidating}.
    \item \textbf{Stochastic Runge-Kutta (SRK)}: We utilize two high-order schemes derived by R\"{o}\ss{}ler. \textbf{SRK2} \citep{rossler2009second} achieves weak order 2 (and strong order 1 for additive noise) using 2 evaluations per step. \textbf{SRK2S} \citep{rossler2010runge} achieves strong order 1 even for general diagonal multiplicative noise while maintaining weak order 2, requiring 3 evaluations per step.
\end{enumerate}

\subsection{Robustness to Discretization Steps}
\label{app:discretization_details}
To evaluate the numerical robustness of our approach, we conducted an experiment analyzing performance across varying discretization steps. As the number of sampling steps increases, CNS exhibits a monotonic decrease in FID, consistently outperforming the standard SDE baseline. Notably, CNS matches the peak FID of the ODE sampler using less than half the number of steps required by the standard SDE. 

Conversely, due to inherent limitations in the pre-trained model's underlying framework, the deterministic ODE sampler fails to maintain a monotonic improvement at high step counts; thus, we omit its results beyond the standard 250 steps. While CNS significantly accelerates stochastic convergence, it still shares the fundamental limitation of SDE solvers, requiring a larger minimum number of discretizations than ODEs to properly integrate the underlying differential equations.

\subsection{Comprehensive Ablation Study}
\label{app:subsec:ablation_study_full}
To empirically validate the theoretical constraints and design choices of the Colored Noise Sampling (CNS) framework, we conduct an extensive ablation study. All ablations are performed on the unguided SiT-XL/2 architecture using 250 Euler sampling steps and evaluated on ImageNet-256 (FID-10K, sFID, and Inception Score). The full quantitative results are detailed in Tab.~\ref{tab:ablation_studies_appendix}. Below, we formalize the methodology and theoretical implications for each ablation category.

\subsubsection{Global Energy Scaling (Validating the Variance Constraint)} 
\begin{wraptable}{r}{0.55\textwidth}
  \vspace{-1em} 
  \centering
  \caption{\textbf{Comprehensive Ablation Studies.} Evaluated on FID-10K ($\downarrow$), sFID ($\downarrow$), and IS ($\uparrow$). We isolate the specific effects of the CNS formulation against alternative baselines, partial noise corruption, temporal schedule permutations, global energy scaling, and multifractional Brownian Motion (mBm) schedules.}
  \label{tab:ablation_studies_appendix}
  \resizebox{\linewidth}{!}{
  \begin{tabular}{l c c c}
    \toprule
    \textbf{Sampling Scenario} & \textbf{FID-10K} $\downarrow$ & \textbf{sFID} $\downarrow$ & \textbf{IS} $\uparrow$ \\
    \midrule
    \textbf{CNS (Ours)} & \textbf{9.61} & \textbf{18.17} & \textbf{143.20} \\
    \midrule
    \rowcolor[gray]{0.95} \multicolumn{4}{l}{\textit{Baselines}} \\
    Deterministic ODE & 17.05 & 23.57 & 98.43 \\
    Standard white-noise SDE & 11.82 & 19.15 & 107.75 \\
    \midrule
    \rowcolor[gray]{0.95} \multicolumn{4}{l}{\textit{Temporal Partial Corruption}} \\
    25\% White Noise & 10.47 & 18.95 & 137.91 \\
    50\% White Noise & 10.64 & 19.08 & 136.36 \\
    50\% Random Unit-Energy Spectrum & 11.28 & 19.21 & 134.81 \\
    100\% Random Unit-Energy Spectrum & 12.26 & 19.80 & 127.09 \\
    \midrule
    \rowcolor[gray]{0.95} \multicolumn{4}{l}{\textit{Temporal Schedule Permutations}} \\
    Constant Spectrum (Temporal Mean) & 10.53 & 19.14 & 137.63 \\
    Shuffled Schedule & 10.46 & 19.03 & 138.02 \\
    Inverted Schedule (Reverse Time) & 10.50 & 18.91 & 137.46 \\
    \midrule
    \rowcolor[gray]{0.95} \multicolumn{4}{l}{\textit{Global Energy Scaling (Variance Constraint Violation)}} \\
    Scale 0.50 (Half Energy) & 106.82 & 278.69 & 9.83 \\
    Scale 0.75 & 53.29 & 161.09 & 34.11 \\
    Scale 0.90 & 16.17 & 42.03 & 111.29 \\
    Scale 1.01 & 11.37 & 20.21 & 133.84 \\
    Scale 1.05 & 20.46 & 29.73 & 96.17 \\
    Scale 1.10 & 50.63 & 49.70 & 46.96 \\
    Scale 1.25 & 171.12 & 91.28 & 5.70 \\
    Scale 1.50 & 256.56 & 134.67 & 2.12 \\
    Scale 2.00 (Double Energy) & 327.45 & 198.37 & 1.55 \\
    \midrule
    \rowcolor[gray]{0.95} \multicolumn{4}{l}{\textit{Multifractional Brownian Motion (mBm) Variants}} \\
    White $\rightarrow$ Blue ($H: 0.5 \rightarrow 0.1$) & 13.46 & 25.16 & 122.49 \\
    White $\rightarrow$ Blue ($H: 0.5 \rightarrow 0.25$) & 11.88 & 19.62 & 130.22 \\
    Red $\rightarrow$ Blue ($H: 0.9 \rightarrow 0.1$) & 266.61 & 224.77 & 2.97 \\
    \bottomrule
  \end{tabular}
  }
  \vspace{-3em} 
\end{wraptable}
In App.~\ref{app:subsec:variance_conservation}, we mathematically established the necessity of the global variance-conservation constraint ($\frac{1}{D}\sum \beta_f^2 = 1$). To empirically prove this, we uniformly scaled the total injected energy budget by factors ranging from $0.50$ to $2.00$. The results are stark: any deviation from a tight neighborhood of the unit-energy budget drastically degrades generation fidelity. Scaling the energy down (e.g., $0.90$) starves the SDE of the necessary stochastic exploration required to correct numerical drift, pulling performance toward the deterministic ODE baseline. Conversely, scaling the energy up by even $5\%$ ($1.05$) destabilizes the process. The excessive injected It\^o heat easily overpowers the restorative gradient of the score network, rapidly pushing the latent states out-of-distribution and destroying semantic coherence.

\subsubsection{Spectral Perturbation and Partial Corruption} 
To validate the precision of our band-wise allocation, we randomly sampled a predefined ratio (25\%, 50\%, 100\%) of timesteps to inject alternative noise distributions rather than the optimal CNS schedule. These injected distributions were either uniform white noise or random unit-energy spectra (where per-band weights are randomized but strictly maintain the global energy budget). As demonstrated in Tab.~\ref{tab:ablation_studies_appendix}, any corruption of the optimal $\beta(f,t)$ mapping monotonically degrades performance. This confirms that merely ensuring a unit-energy budget is insufficient; the energy must be explicitly routed to the specific frequency bands experiencing the highest structural deficit.

\subsubsection{Temporal Schedule Permutations}
As theorized in App.~\ref{app:correlation_decay_mechanisms}, the state-error correlation decays as structures resolve, meaning the efficiency of energy injection is highly non-stationary. To prove that \textit{when} energy is injected is just as critical as \textit{where}, we dismantled the temporal alignment of the CNS schedule:
\begin{itemize}
    \item \textbf{Constant Spectrum:} We averaged the dynamic CNS matrix across all timesteps to create a single, static colored noise profile.
    \item \textbf{Shuffled Schedule:} We randomly permuted the timesteps of the allocation matrix.
    \item \textbf{Inverted Schedule:} We applied the schedule backwards relative to the generative time.
\end{itemize}
While all three variants maintain the exact same total frequency-wise energy injection as the optimal CNS over the full trajectory, they completely fail to match its FID. By injecting energy at incorrect timesteps—often into bands that are already structurally resolved—they squander the finite variance budget on transient rotational noise, validating the necessity of a dynamic, state-aware schedule.

\subsubsection{Alternative Noise Formulations (mBm)}
Finally, we explored an alternative mathematical formulation for generating non-stationary colored noise: multifractional Brownian Motion (mBm) \citep{peltier1995multifractional}. Unlike standard Wiener processes, mBm introduces a time-varying Hurst parameter, $H(t)$, allowing the stochastic increments to shift continuously between distinct noise colors (e.g., from $H=0.5$ White noise to $H < 0.5$ Blue noise). Formally, this frequency-dependent shift can be understood through the harmonizable representation of mBm:
\begin{equation}
B_{H(t)}(t) = \frac{1}{C(H(t))} \int_{\mathbb{R}} \frac{e^{i t \omega} - 1}{|\omega|^{H(t) + 1/2}} d\tilde{W}(\omega)
\end{equation}
where $\omega$ denotes frequency, $d\tilde{W}(\omega)$ is the complex Wiener measure, and $C(H(t))$ is a normalization constant. In this form, the exponent $H(t) + 1/2$ directly governs the spectral density, mathematically enforcing how the noise color evolves over time.
While theoretically elegant, mBm dictates a rigid, parameterized shift across the entire spectrum. It lacks the highly granular, per-band structural awareness provided by the $\gamma$-matrix. Consequently, even meticulously tuned mBm schedules (e.g., White $\to$ Blue, $H: 0.5 \to 0.25$) fall short of performance of the proposed CNS framework, although optimized mBm configurations can still offer marginal improvements over the standard SDE.

\subsection{Generalization to Alternative Noise Training}
\label{app:bndm_details}
To demonstrate that our framework is orthogonal to alternative noise training paradigms, we evaluated CNS on the official pre-trained IADB models provided by the authors of BNDM \cite{huang2024blue}. Because these BNDM experiments rely on a unique, temporally evolving noise distribution, we first derived and implemented a custom SDE sampler strictly tailored to their specific forward process. Despite the alternative training objective, we observed that these models still exhibit a pronounced spectral bias. We empirically tracked this bias to compute the corresponding structural progression $\gamma(f,t)$-matrix and integrated our CNS method on top of the custom SDE solver. By dynamically shaping the injected noise spectrum according to this matrix, CNS achieved significant generation improvements across two evaluated $64 \times 64$ datasets using the standard provided test batches.

\subsection{Additional Visual Comparisons}
\label{sec:appendix_Visual_compare}

\begin{figure}[h]
    \centering
    \begin{minipage}{\textwidth}
        \centering
        \begin{minipage}{0.15\textwidth}\centering \textbf{CNS (Ours)}\end{minipage}\hfill
        \begin{minipage}{0.15\textwidth}\centering \textbf{SDE}\end{minipage}\hfill
        \begin{minipage}{0.15\textwidth}\centering \textbf{ODE}\end{minipage}\hfill
        \hspace{0.02\textwidth} 
        \begin{minipage}{0.15\textwidth}\centering \textbf{CNS (Ours)}\end{minipage}\hfill
        \begin{minipage}{0.15\textwidth}\centering \textbf{SDE}\end{minipage}\hfill
        \begin{minipage}{0.15\textwidth}\centering \textbf{ODE}\end{minipage}
        \vspace{0.5em}
    \end{minipage}
    
    \begin{minipage}{0.48\textwidth}
        \begin{minipage}{\textwidth}
            \centering
            \begin{subfigure}{0.32\textwidth}
                \includegraphics[width=\linewidth]{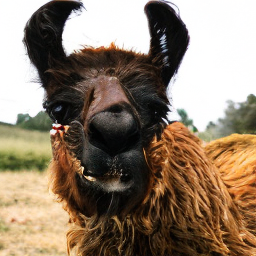}
            \end{subfigure}\hfill
            \begin{subfigure}{0.32\textwidth}
                \includegraphics[width=\linewidth]{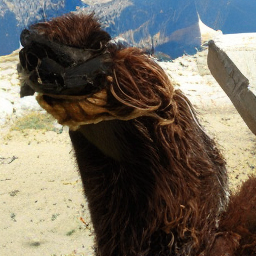}
            \end{subfigure}\hfill
            \begin{subfigure}{0.32\textwidth}
                \includegraphics[width=\linewidth]{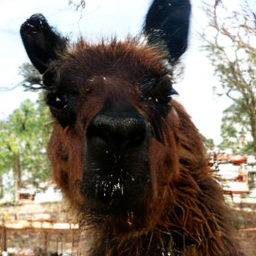}
            \end{subfigure}
            \caption*{Class: Llama}
            \vspace{1em} 
        \end{minipage}
    \end{minipage}\hfill
    \begin{minipage}{0.48\textwidth}
        \begin{minipage}{\textwidth}
            \centering
            \begin{subfigure}{0.32\textwidth}
                \includegraphics[width=\linewidth]{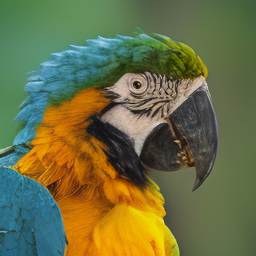}
            \end{subfigure}\hfill
            \begin{subfigure}{0.32\textwidth}
                \includegraphics[width=\linewidth]{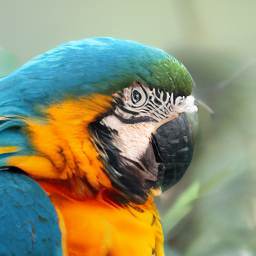}
            \end{subfigure}\hfill
            \begin{subfigure}{0.32\textwidth}
                \includegraphics[width=\linewidth]{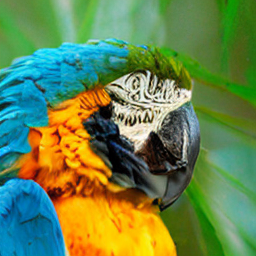}
            \end{subfigure}
            \caption*{Class: Macaw}
            \vspace{1em} 
        \end{minipage}
    \end{minipage}
    \vspace{1em}
    
    \begin{minipage}{0.48\textwidth}
        \begin{minipage}{\textwidth}
            \centering
            \begin{subfigure}{0.32\textwidth}
                \includegraphics[width=\linewidth]{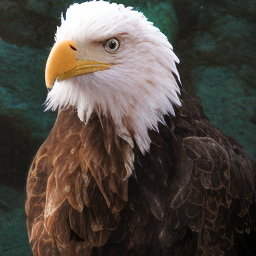}
            \end{subfigure}\hfill
            \begin{subfigure}{0.32\textwidth}
                \includegraphics[width=\linewidth]{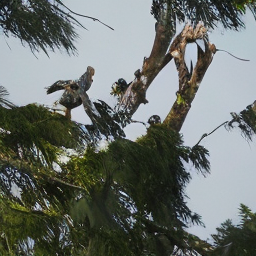}
            \end{subfigure}\hfill
            \begin{subfigure}{0.32\textwidth}
                \includegraphics[width=\linewidth]{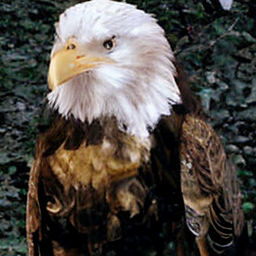}
            \end{subfigure}
            \caption*{Class: Bald Eagle}
            \vspace{1em} 
        \end{minipage}
    \end{minipage}\hfill
    \begin{minipage}{0.48\textwidth}
        \begin{minipage}{\textwidth}
            \centering
            \begin{subfigure}{0.32\textwidth}
                \includegraphics[width=\linewidth]{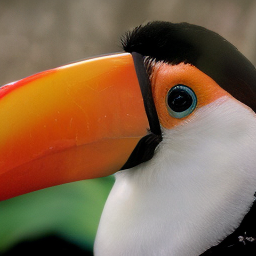}
            \end{subfigure}\hfill
            \begin{subfigure}{0.32\textwidth}
                \includegraphics[width=\linewidth]{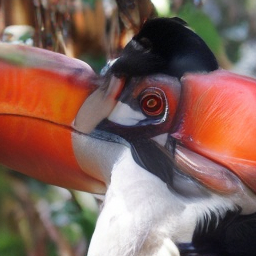}
            \end{subfigure}\hfill
            \begin{subfigure}{0.32\textwidth}
                \includegraphics[width=\linewidth]{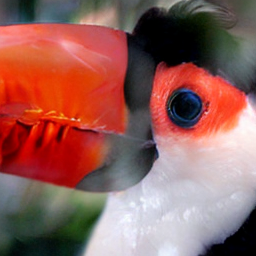}
            \end{subfigure}
            \caption*{Class: Toucan}
            \vspace{1em} 
        \end{minipage}
    \end{minipage}
    \vspace{1em}
    
    \begin{minipage}{0.48\textwidth}
        \begin{minipage}{\textwidth}
            \centering
            \begin{subfigure}{0.32\textwidth}
                \includegraphics[width=\linewidth]{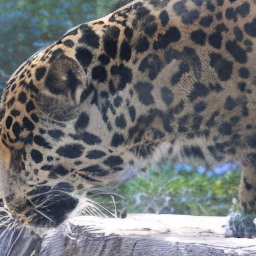}
            \end{subfigure}\hfill
            \begin{subfigure}{0.32\textwidth}
                \includegraphics[width=\linewidth]{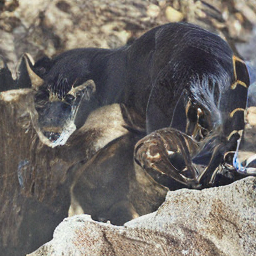}
            \end{subfigure}\hfill
            \begin{subfigure}{0.32\textwidth}
                \includegraphics[width=\linewidth]{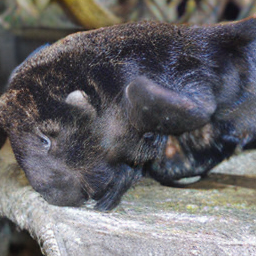}
            \end{subfigure}
            \caption*{Class: Jaguar}
            \vspace{1em} 
        \end{minipage}
    \end{minipage}\hfill
    \begin{minipage}{0.48\textwidth}
        \begin{minipage}{\textwidth}
            \centering
            \begin{subfigure}{0.32\textwidth}
                \includegraphics[width=\linewidth]{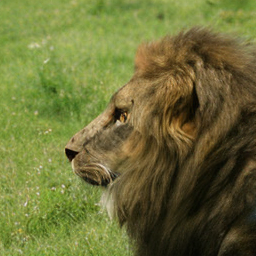}
            \end{subfigure}\hfill
            \begin{subfigure}{0.32\textwidth}
                \includegraphics[width=\linewidth]{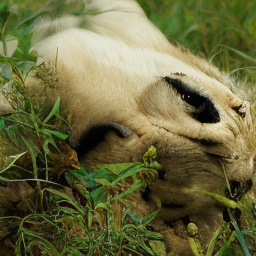}
            \end{subfigure}\hfill
            \begin{subfigure}{0.32\textwidth}
                \includegraphics[width=\linewidth]{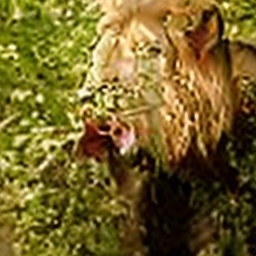}
            \end{subfigure}
            \caption*{Class: Lion}
            \vspace{1em} 
        \end{minipage}
    \end{minipage}
    \vspace{1em}
    
    \begin{minipage}{0.48\textwidth}
        \begin{minipage}{\textwidth}
            \centering
            \begin{subfigure}{0.32\textwidth}
                \includegraphics[width=\linewidth]{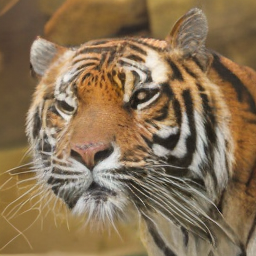}
            \end{subfigure}\hfill
            \begin{subfigure}{0.32\textwidth}
                \includegraphics[width=\linewidth]{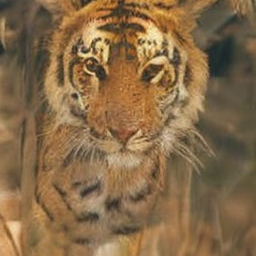}
            \end{subfigure}\hfill
            \begin{subfigure}{0.32\textwidth}
                \includegraphics[width=\linewidth]{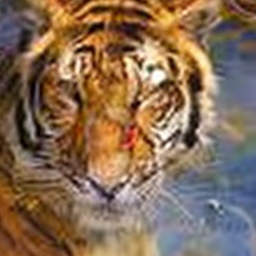}
            \end{subfigure}
            \caption*{Class: Tiger}
            \vspace{1em} 
        \end{minipage}
    \end{minipage}\hfill
    \begin{minipage}{0.48\textwidth}
        \begin{minipage}{\textwidth}
            \centering
            \begin{subfigure}{0.32\textwidth}
                \includegraphics[width=\linewidth]{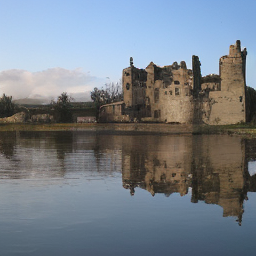}
            \end{subfigure}\hfill
            \begin{subfigure}{0.32\textwidth}
                \includegraphics[width=\linewidth]{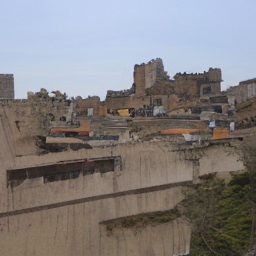}
            \end{subfigure}\hfill
            \begin{subfigure}{0.32\textwidth}
                \includegraphics[width=\linewidth]{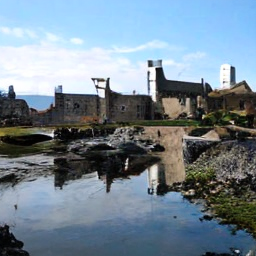}
            \end{subfigure}
            \caption*{Class: Castle}
            \vspace{1em} 
        \end{minipage}
    \end{minipage}

\caption{Visual comparison of samples generated using ODE, standard SDE, and our proposed CNS framework (without CFG). All images in a given triplet were generated using the same seed.}
    \label{fig:comparison_grid_1}
\end{figure}

\begin{figure}[h] 
    \centering
    \begin{minipage}{\textwidth}
        \centering
        \begin{minipage}{0.15\textwidth}\centering \textbf{CNS (Ours)}\end{minipage}\hfill
        \begin{minipage}{0.15\textwidth}\centering \textbf{SDE}\end{minipage}\hfill
        \begin{minipage}{0.15\textwidth}\centering \textbf{ODE}\end{minipage}\hfill
        \hspace{0.02\textwidth} 
        \begin{minipage}{0.15\textwidth}\centering \textbf{CNS (Ours)}\end{minipage}\hfill
        \begin{minipage}{0.15\textwidth}\centering \textbf{SDE}\end{minipage}\hfill
        \begin{minipage}{0.15\textwidth}\centering \textbf{ODE}\end{minipage}
        \vspace{0.5em}
    \end{minipage}
    
    \begin{minipage}{0.48\textwidth}
        \begin{minipage}{\textwidth}
            \centering
            \begin{subfigure}{0.32\textwidth}
                \includegraphics[width=\linewidth]{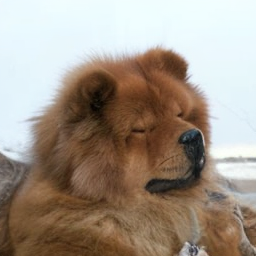}
            \end{subfigure}\hfill
            \begin{subfigure}{0.32\textwidth}
                \includegraphics[width=\linewidth]{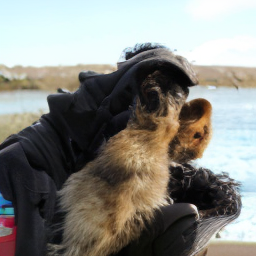}
            \end{subfigure}\hfill
            \begin{subfigure}{0.32\textwidth}
                \includegraphics[width=\linewidth]{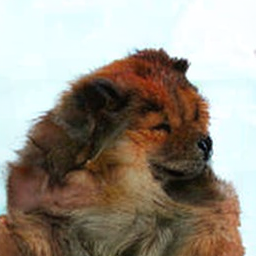}
            \end{subfigure}
            \caption*{Class: Chow}
            \vspace{1em} 
        \end{minipage}
    \end{minipage}\hfill
    \begin{minipage}{0.48\textwidth}
        \begin{minipage}{\textwidth}
            \centering
            \begin{subfigure}{0.32\textwidth}
                \includegraphics[width=\linewidth]{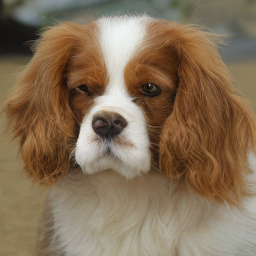}
            \end{subfigure}\hfill
            \begin{subfigure}{0.32\textwidth}
                \includegraphics[width=\linewidth]{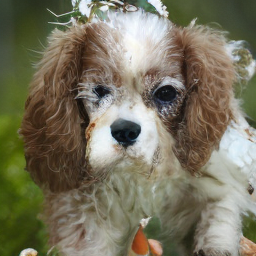}
            \end{subfigure}\hfill
            \begin{subfigure}{0.32\textwidth}
                \includegraphics[width=\linewidth]{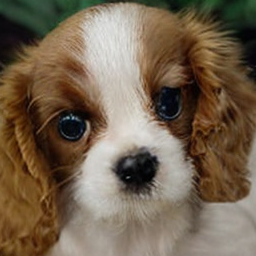}
            \end{subfigure}
            \caption*{Class: Blenheim Spaniel}
            \vspace{1em} 
        \end{minipage}
    \end{minipage}
    \vspace{1em}
    
    \begin{minipage}{0.48\textwidth}
        \begin{minipage}{\textwidth}
            \centering
            \begin{subfigure}{0.32\textwidth}
                \includegraphics[width=\linewidth]{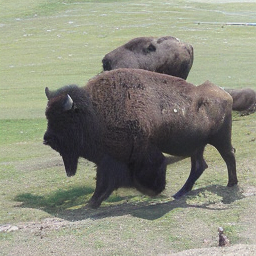}
            \end{subfigure}\hfill
            \begin{subfigure}{0.32\textwidth}
                \includegraphics[width=\linewidth]{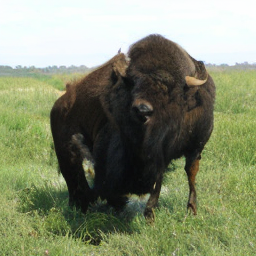}
            \end{subfigure}\hfill
            \begin{subfigure}{0.32\textwidth}
                \includegraphics[width=\linewidth]{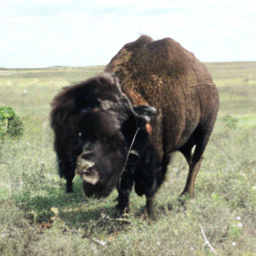}
            \end{subfigure}
            \caption*{Class: Bison}
            \vspace{1em} 
        \end{minipage}
    \end{minipage}\hfill
    \begin{minipage}{0.48\textwidth}
        \begin{minipage}{\textwidth}
            \centering
            \begin{subfigure}{0.32\textwidth}
                \includegraphics[width=\linewidth]{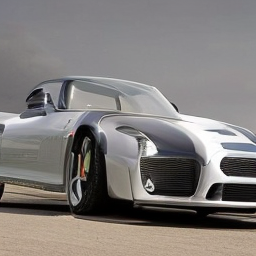}
            \end{subfigure}\hfill
            \begin{subfigure}{0.32\textwidth}
                \includegraphics[width=\linewidth]{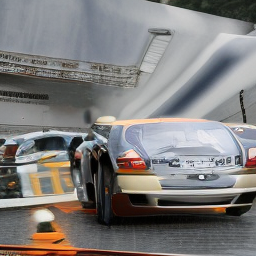}
            \end{subfigure}\hfill
            \begin{subfigure}{0.32\textwidth}
                \includegraphics[width=\linewidth]{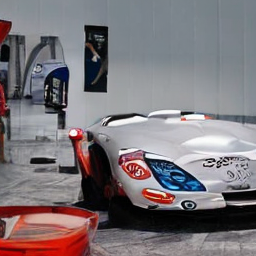}
            \end{subfigure}
            \caption*{Class: Sports Car}
            \vspace{1em} 
        \end{minipage}
    \end{minipage}
    \vspace{1em}
    
    \begin{minipage}{0.48\textwidth}
        \begin{minipage}{\textwidth}
            \centering
            \begin{subfigure}{0.32\textwidth}
                \includegraphics[width=\linewidth]{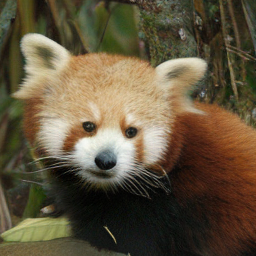}
            \end{subfigure}\hfill
            \begin{subfigure}{0.32\textwidth}
                \includegraphics[width=\linewidth]{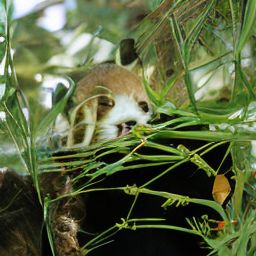}
            \end{subfigure}\hfill
            \begin{subfigure}{0.32\textwidth}
                \includegraphics[width=\linewidth]{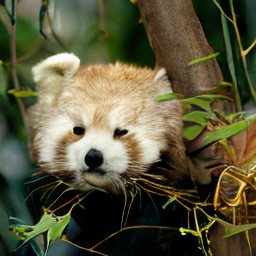}
            \end{subfigure}
            \caption*{Class: Lesser Panda}
            \vspace{1em} 
        \end{minipage}
    \end{minipage}\hfill
    \begin{minipage}{0.48\textwidth}
        \begin{minipage}{\textwidth}
            \centering
            \begin{subfigure}{0.32\textwidth}
                \includegraphics[width=\linewidth]{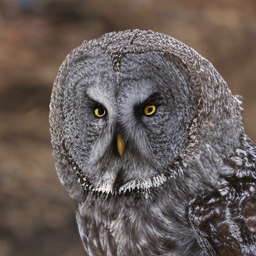}
            \end{subfigure}\hfill
            \begin{subfigure}{0.32\textwidth}
                \includegraphics[width=\linewidth]{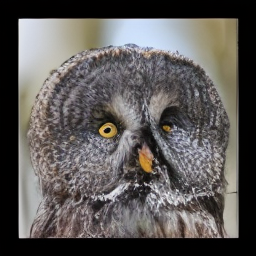}
            \end{subfigure}\hfill
            \begin{subfigure}{0.32\textwidth}
                \includegraphics[width=\linewidth]{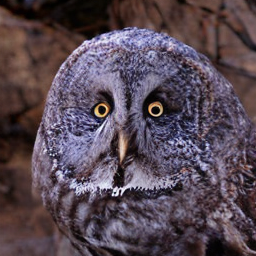}
            \end{subfigure}
            \caption*{Class: Great Grey Owl}
            \vspace{1em} 
        \end{minipage}
    \end{minipage}
    \vspace{1em}
    
    \begin{minipage}{0.48\textwidth}
        \begin{minipage}{\textwidth}
            \centering
            \begin{subfigure}{0.32\textwidth}
                \includegraphics[width=\linewidth]{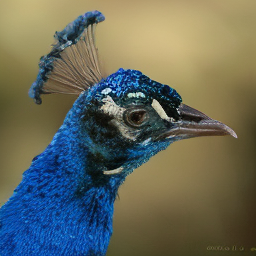}
            \end{subfigure}\hfill
            \begin{subfigure}{0.32\textwidth}
                \includegraphics[width=\linewidth]{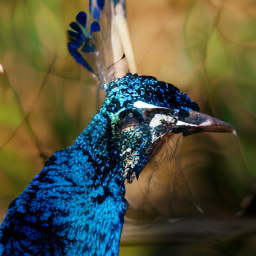}
            \end{subfigure}\hfill
            \begin{subfigure}{0.32\textwidth}
                \includegraphics[width=\linewidth]{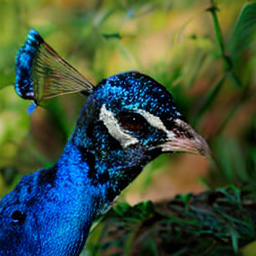}
            \end{subfigure}
            \caption*{Class: Peacock}
            \vspace{1em} 
        \end{minipage}
    \end{minipage}\hfill
    \begin{minipage}{0.48\textwidth}
        \begin{minipage}{\textwidth}
            \centering
            \begin{subfigure}{0.32\textwidth}
                \includegraphics[width=\linewidth]{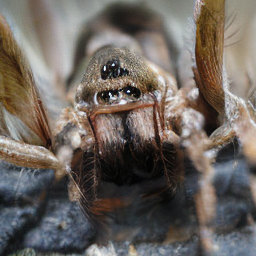}
            \end{subfigure}\hfill
            \begin{subfigure}{0.32\textwidth}
                \includegraphics[width=\linewidth]{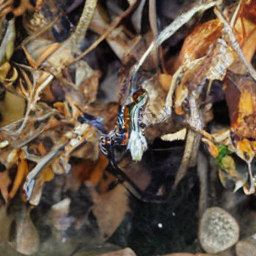}
            \end{subfigure}\hfill
            \begin{subfigure}{0.32\textwidth}
                \includegraphics[width=\linewidth]{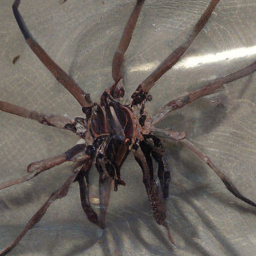}
            \end{subfigure}
            \caption*{Class: Wolf Spider}
            \vspace{1em} 
        \end{minipage}
    \end{minipage}

    \caption{Visual comparison of samples generated using ODE, standard SDE, and our proposed CNS framework (without CFG). All images in a given triplet were generated using the same seed.}
    \label{fig:comparison_grid_2}
\end{figure}

In this section, we provide extended qualitative results demonstrating the efficacy of our proposed Colored Noise Sampling (CNS) framework compared to standard deterministic (ODE) and stochastic (SDE) baselines. Fig.~\ref{fig:comparison_grid_1} and Fig.~\ref{fig:comparison_grid_2} present generation samples across a diverse set of ImageNet classes. To ensure a fair and isolated evaluation of the sampling dynamics, all images within a given row are generated using the exact same noise realizations and class prompt. 

As theoretically established earlier, standard uniform white-noise SDEs frequently struggle to resolve high-frequency spatial structures, often yielding blurry or structurally degraded textures. Conversely, while deterministic ODEs preserve structure, they suffer from accumulation errors that lead to over-smoothed, artificial appearances. By dynamically routing stochastic energy to structurally unresolved frequency bands, CNS consistently bridges this gap, yielding sharper fine details (e.g., fur, feathers, and foliage) and a more globally coherent output that better aligns with the true data manifold.

\newpage

\end{document}